\documentclass[pdflatex,sn-mathphys-num]{sn-jnl}

\usepackage{graphicx}
\usepackage{adjustbox}
\usepackage{rotating}
\usepackage{amsmath,amssymb,amsfonts}
\usepackage{array}
\usepackage{booktabs}
\usepackage{multirow}
\usepackage{tabularx}
\usepackage{siunitx}
\usepackage{textcomp}
\usepackage{verbatim}
\usepackage{afterpage}
\usepackage{placeins}
\usepackage{float}
\usepackage{newunicodechar}

\newunicodechar{≈}{\ensuremath{\approx}}
\newunicodechar{₂}{\ensuremath{{}_2}}

\begin{document}

\title[Enrichment-oriented screening of perovskite devices]
{Enrichment-Oriented Virtual Screening of Perovskite Solar Cells
through Reinforcement Learning}

\author[1]{\fnm{Yezeyuan} \sur{Zheng}}
\email{yzheng@hkqai.hk}

\author[1,2]{\fnm{Ziyang} \sur{Hu}}
\email{hzy@yangtze.hku.hk}

\author[6]{\fnm{Maojun} \sur{Sun}}
\email{mj.sun@connect.polyu.hk}

\author[6]{\fnm{Ruijian} \sur{Han}}
\email{ruijian.han@polyu.edu.hk}

\author[6]{\fnm{Jian} \sur{Huang}}
\email{j.huang@polyu.edu.hk}

\author[1,2,3,4,5]{\fnm{Guanhua} \sur{Chen}}
\email{ghc@everest.hku.hk}

\author[1,2]{\fnm{Jie} \sur{Liu}}
\email{jie5478@hku.hk}

\affil[1]{Hong Kong Quantum AI Lab Limited, Pak Shek Kok, Hong Kong SAR, China}
\affil[2]{Department of Chemistry, The University of Hong Kong, Pokfulam, Hong Kong SAR, China}
\affil[3]{MattVerse Limited, Pak Shek Kok, Hong Kong SAR, China}
\affil[4]{QuantumFabless Limited, Pak Shek Kok, Hong Kong SAR, China}
\affil[5]{Lixin Technology (Shenzhen) Co. Ltd., Shenzhen, Guangdong, China}

\affil[6]{Department of Data Science and AI, and Applied Mathematics, The Hong Kong Polytechnic University, Hung Hom, Kowloon, Hong Kong SAR, China}
\abstract{
Experimental optimization of perovskite solar cells is constrained by combinatorial diversity in absorber compositions, device layers, and processing, which makes virtual screening attractive for prioritizing recipes before fabrication. We develop a surrogate-guided reinforcement-learning framework that frames recipe generation as virtual screening, enriching candidate shortlists near high-performing experimental neighbors. A frozen Tabular Prior-data Fitted Network (TabPFN) guides recipe generation, while a reference pool of 890 measured devices enables neighborhood-based assessment. Five-fold evaluation on 1,846 devices shows top-ten selection raises mean power conversion efficiency (PCE) from 19.68\% to 23.74\%. Three learned policies are compared against six baselines under 3,000 surrogate evaluations per run. In top-100 shortlists, Proximal policy optimization (PPO) and recurrent PPO (RPPO) achieve neighborhood-supported hit rates of 23\% and 24\% (7.9- and 8.3-fold Pure Random's 2.9\%), versus 10\% (3.4-fold) for the baselines with the highest supported hit rate (L-SHADE and Ax BO, tied) at $K = 100$. Gains coexist with weak-support fractions. Of 9,000 recipes, 65.4\% match a composition family at similarity 0.80, falling to 11.7\% at 0.90. Results support enrichment-oriented prioritization but not experimental performance, seed robustness, or matched-budget advantage.
}


\maketitle

%

\section{Introduction}

Perovskite solar cells combine tunable optoelectronic properties, strong optical absorption, and compatibility with solution-based
fabrication. Their development nevertheless requires navigating a large design space spanning absorber composition, charge-transport
layers, electrode materials, and fabrication procedures.
Interactions among these variables make experimental optimization expensive: improving one component does not necessarily improve the complete device, and exhaustive exploration of their combinations
is impractical. Effective computational screening could therefore help allocate experimental effort to a manageable subset of relatively more promising configurations.

Machine learning offers a route to such virtual screening. The Perovskite Database Project~\cite{jacobsson2022perovskite}
provides a substantial experimental foundation for performance
prediction and candidate screening. Previous studies have used models trained on this database to investigate Sn-based devices and
richer composition--architecture representations
\cite{rumman2023sn,kusuma2025stability}.
However, missing synthesis information, inconsistent reporting,
duplicate measurements, and differences between experimental
studies limit predictive reliability. These limitations become
particularly important during optimization, when a search
algorithm may exploit errors in a predictive model rather than
identify genuinely improved devices.

An enrichment-oriented screening perspective separates the
optimization objective from the evidence used to assess a shortlist. This mirrors a virtual-screening workflow: a policy generates a virtual library of recipes, a surrogate scores and ranks them, and the top-ranked shortlist is assessed against separately constructed experimental references pool. Rather than interpreting the largest predicted power conversion efficiency (PCE)
as an achievable device efficiency, the immediate objective is
to increase the fraction of shortlisted configurations with
support from relevant experimental records. Surrogate scores
can guide candidate generation, while reference-based analyses
provide complementary information about the resulting proposals.
Such analyses remain computational assessments and actual fabrication and measurement should follow that.

Reinforcement learning (RL) provides one approach to learning
reusable candidate-generation policies over combinatorial
design spaces. Applications to reaction optimization, processing
recipes, and compositional design demonstrate the flexibility
of policy-based search
\cite{zhou2017dro,rajak2021cvd,karpovich2024inverse,xian2024alloys}.
Related work has also combined surrogate modeling and
reinforcement learning for photovoltaic inverse design
\cite{zhang2026doubleabsorber}.
For recipe screening, the relevant opportunity is to learn proposal distributions that repeatedly generate useful candidates, so that RL-based optimization of the surrogate serves as the candidate-generation stage of a virtual-screening pipeline, rather than necessarily to perform long-horizon synthesis planning.

Here, we investigate whether RL-based recipe generation, used as a virtual-screening strategy, can enrich perovskite-device shortlists in configurations
supported by high-performing experimental neighborhoods.
We assess three complementary questions: whether the surrogate
and neighborhood scores contain useful ranking information on
measured devices; whether policy-generated shortlists show
greater high-PCE neighborhood support than conventional
optimization methods; and how the resulting proposals relate
to composition families in separately retrieved literature.
Throughout, we distinguish surrogate optimization,
experimental-reference support, and experimentally measured
device performance.

\section{Results}
\label{sec:results}

\subsection{High-dimensional representation and surrogate-guided virtual
screening of perovskite devices}
\label{sec:framework_overview}

\begin{figure}[htbp]
    \centering
    \includegraphics[width=\textwidth]
    {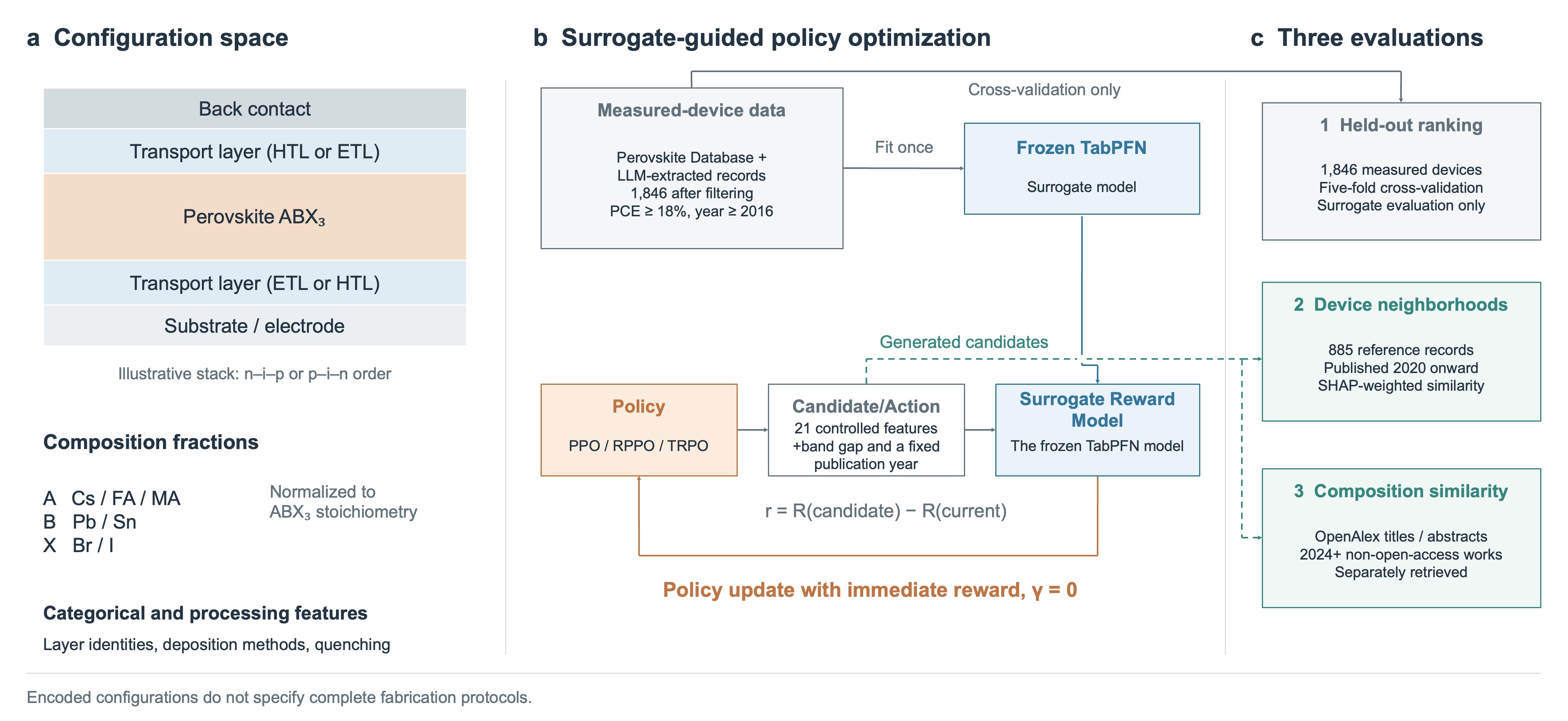}
    \caption{\textbf{Overview of the configuration space, surrogate-guided policy optimization, virtual-screening workflow, and evaluation framework for perovskite solar-cell design.}
    \textbf{(a)} Device configurations encode layer identities, ABX$_3$ composition fractions normalized to stoichiometry, and processing features for n--i--p or p--i--n architectures; these representations do not specify complete fabrication protocols.
    \textbf{(b)} A TabPFN surrogate is fitted on a 1{,}292-record training partition
    drawn from a 1{,}846-record high-PCE subset of the Perovskite Database
    and large language model (LLM)-extracted literature (PCE \(\geq 18\%\), publication year
    \(\geq 2016\); stratified 70:30 split, seed 50), and is then frozen
    for policy optimization. Proximal policy optimization (PPO), recurrent PPO (RPPO), or trust region policy optimization (TRPO) policies propose candidates defined by 21 controlled features, together with band gap and a fixed publication year. Policy updates use the immediate change in surrogate score, $r = R(\mathrm{candidate}) - R(\mathrm{current})$ (where candidate corresponds to \(s_{t+1}\) and current corresponds to \(s_{t}\) as shown in the methods, with discount factor $\gamma = 0$).
    \textbf{(c)} Three complementary evaluations assess held-out surrogate ranking through five-fold cross-validation, generated-candidate similarity to 890 reference devices published from 2020 onward using SHapley Additive exPlanations (SHAP)-weighted device neighborhoods, and composition similarity to separately retrieved OpenAlex titles and abstracts of non-open-access works published from 2024 onward.}
    \label{fig:framework}
\end{figure}

We developed a computational virtual-screening workflow comprising recipe representation, RL-driven candidate generation, surrogate-based ranking and shortlisting, and experimental-reference assessment
(Figure~\ref{fig:framework}).
Each candidate is represented by 21 controlled variables
encoding layer materials, deposition procedures, absorber
composition, and a quenching flag. Normalization enforces the
represented $\mathrm{ABX_3}$ stoichiometry. A
composition-derived band gap and a fixed publication year
complete the 23-feature input to a frozen Tabular Prior-data Fitted Network (TabPFN) reward model.

The proposal policies optimize the surrogate score, whereas
the subsequent neighborhood analysis assesses support from
measured reference devices. The neighborhood criterion is
not used as the policy reward. A separate composition-family
analysis characterizes the compositional regions occupied
by the generated candidates.

This separation is central to the screening interpretation:
a high surrogate score indicates successful optimization of
the fitted model, while a high-PCE neighborhood indicates
support from experimentally reported analogues under the
chosen similarity metric. 

\subsection{Validation of surrogate ranking and
experimental-neighborhood signals}
\label{sec:reward_model_results}

We first examined whether the two screening signals contain
information relevant to prioritizing experimentally measured
devices. Surrogate performance was evaluated by five-fold
cross-validation on 1,846 deduplicated high-PCE records,
using fold-specific preprocessing and band-gap imputation.
The mean fold-level performance was
$R^2=0.414\pm0.077$,
MAE $=0.867\pm0.021$ percentage points, and
RMSE $=1.118\pm0.042$ percentage points
(Table~\ref{tab:reward}).
These results indicate moderate predictive accuracy. We treat the five-fold cross-validation as the primary evaluation, since
it tests every record out of sample, averages over partitions, and
reports variability across folds. The frozen reward model used in
optimization, evaluated on its single held-out split ($n=554$), gave
$\mathrm{RMSE}=1.156$ percentage points. As a single split, it provides a
complementary check on the deployed model but a noisier estimate.

\begin{figure}[htbp]
    \centering
    \includegraphics[width=\textwidth]
    {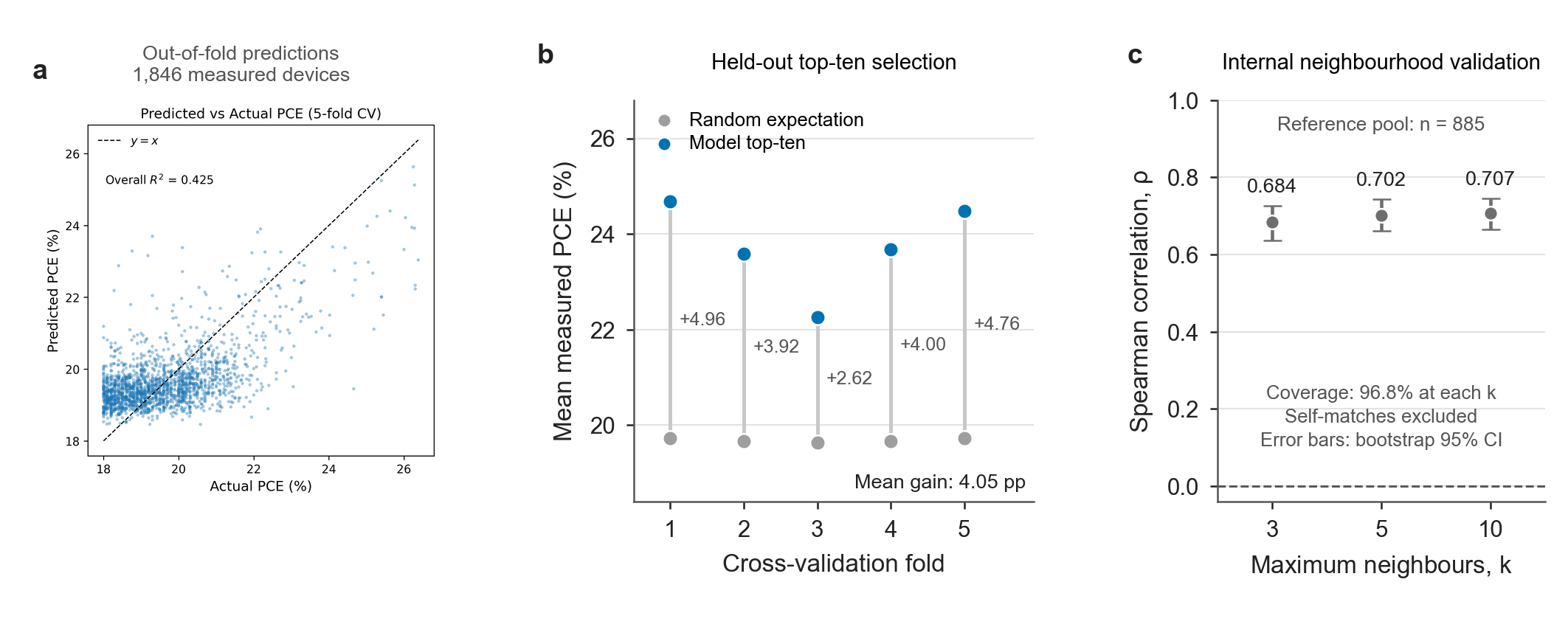}
    \caption{
    Validation of surrogate ranking and the
    reference-neighborhood signal.
    (a) Predicted versus measured PCE for all 1,846 devices, using out-of-fold predictions from five-fold stratified cross-validation (each point was predicted by a model that did not see that device).
    (b) Mean measured PCE of the ten highest-scoring
    devices in each validation fold compared with the
    random-selection expectation, defined as the mean
    measured PCE of that fold. Paired-point annotations
    show selection gains in percentage points (pp);
    the mean gain across folds is 4.05~pp.
    (c) Spearman correlation between measured PCE and
    neighborhood-mean PCE within the 890-device reference
    pool for maximum neighborhood sizes of $k=3$, 5,
    and 10. Self-matches are excluded and eligible
    neighbors have similarity $\geq0.50$.
    Error bars show reported bootstrap 95\% confidence
    intervals; coverage is 96.4\% at each setting.
    Panel (c) assesses the neighborhood signal internally
    within the reference pool, not the experimental
    performance of generated candidates.
    Detailed ranking results, residual analysis, and
    neighborhood-validation controls are provided in
    Appendices~\ref{app:surrogate-ranking},
    \ref{app:residual_analysis}, and
    \ref{app:signal_validation}, respectively.
    }
    \label{fig:pred_vs_actual}
\end{figure}

Ranking performance was more directly relevant to the
screening objective. Selecting the ten highest-scoring
devices within each validation fold gave a mean measured
PCE of 23.74\%, compared with 19.68\% under uniform random
selection from the same folds. The mean gain was therefore
4.05 percentage points and was positive in every fold.
The mean within-fold Spearman correlation was
$0.436\pm0.037$
(Figure~\ref{fig:pred_vs_actual};
Appendix~\ref{app:surrogate-ranking}).

Table~\ref{tab:reward} also reports full-dataset performance
and comparisons with previously reported modeling approaches. Our model performed better on the high-PCE subset, but worse on the full cleaned dataset comparing to the baselines.
Because the datasets and preprocessing protocols differ, these comparisons are indicative. Residual analysis showed
increasing underprediction at higher measured PCE
(Appendix~\ref{app:residual_analysis}).

We next evaluated the neighborhood signal within the 890-record experimental reference pool. Each reference device was compared with its eligible neighbors while excluding self-matches. Neighborhood-mean PCE was correlated
with the device's measured PCE, with Spearman correlations of 0.682, 0.699, and 0.707 for maximum neighborhood sizes
of $k = 3$, 5, and 10, respectively. Coverage was 96.4\% at each setting. PCE-label permutation and random-neighbor controls supported the informativeness of similarity-based
neighbor selection (Figure~\ref{fig:pred_vs_actual}c;
Appendix~\ref{app:signal_validation}).

Together, these analyses support two distinct uses of the available experimental information: surrogate scores guide recipe generation, while neighborhood scores characterize reference support for the resulting candidates.

\begin{table}[htbp]
    \centering
    \caption{
    Reward-model performance.
    Values with uncertainties are reported as mean $\pm$ standard
    deviation across five cross-validation folds.
    MAE and RMSE are expressed in percentage points of PCE.
    Full-dataset baseline results are taken from the original
    publications and use their respective data preparation and
    evaluation protocols.
    High-PCE baseline reproductions retain different record subsets
    under their respective preprocessing pipelines.
    These comparisons are indicative.
    }
    \label{tab:reward}
    \small
    \setlength{\tabcolsep}{1pt}
    \renewcommand{\arraystretch}{1.0}
    \begin{tabular*}{\textwidth}{@{\extracolsep{\fill}}l ccc ccc@{}}
        \toprule
        \multirow{2}{*}{\textbf{Model}}
        & \multicolumn{3}{c}{\textbf{Full dataset}}
        & \multicolumn{3}{c}{\textbf{High-PCE subset}} \\
        \cmidrule(lr){2-4}
        \cmidrule(lr){5-7}
        & $R^2$ & MAE & RMSE
        & $R^2$ & MAE & RMSE \\
        \midrule
        Rumman~\cite{rumman2023sn}
        & 0.706 & 1.980 & 2.802
        & $0.201\pm0.034$
        & $1.401\pm0.037$
        & $2.064\pm0.043$ \\
        Kusuma~\cite{kusuma2025stability}
        & 0.751 & 1.901 & 2.868
        & $0.335\pm0.056$
        & $1.274\pm0.030$
        & $1.883\pm0.085$ \\
        Ours
        & $0.526 \pm 0.012$
        & $2.593 \pm 0.050$
        & $3.485 \pm 0.050$
        & $0.414 \pm 0.077$
        & $0.867 \pm 0.021$
        & $1.118 \pm 0.042$ \\
        \bottomrule
    \end{tabular*}
\end{table}

\subsection{Learning recipe-proposal policies and optimizing
the surrogate objective}
\label{sec:rl_results}
\label{sec:baseline_comparison}

We trained PPO, RPPO, and TRPO to propose
recipes using the immediate change in frozen surrogate
score as the reward. Because each action specifies the
complete next encoded recipe and the main experiments use
$\gamma=0$, the decision objective is immediate
next-recipe scoring rather than long-horizon planning.
All three policies improved during the reported training
runs. A discount factor of 0.99 produced substantially
weaker improvement under the evaluated settings.
Training trajectories and the discount-factor analysis are
provided in Appendices~\ref{app:training_diagnostics}
and~\ref{app:gamma}.

We then compared the frozen policies with six optimization baselines: random search; L-SHADE (success-history-based adaptive differential evolution with linear population-size reduction); particle swarm optimization (PSO); the Runge–Kutta optimizer (RUN); covariance matrix adaptation evolution strategy (CMA-ES); and Bayesian optimization implemented in Ax platform (Ax BO) under a common candidate-scoring budget of
3,000 reward-model evaluations per run. Each RL method
was evaluated using three inference seeds of one trained
policy, whereas each baseline used three separately
initialized optimization runs.

The mean best-found surrogate scores were
$34.89\pm0.04$ for RPPO,
$34.50\pm0.35$ for PPO, and
$33.30\pm0.61$ for TRPO.
The non-RL baseline with the highest surrogate score was L-SHADE, with $30.36\pm0.24$
(Table~\ref{tab:convergence}).
The highest surrogate scores (34.89\%, 34.50\% and 33.30\%) exceed or reach the
single-junction Shockley--Queisser limit (approximately 33\%). They should
therefore be interpreted as surrogate extrapolations, which likely reflect
some degree of surrogate exploitation, and not as predicted device
efficiencies. We use them only to rank methods under a common reward.

The comparison establishes that the trained policies effectively optimize the frozen surrogate under the reported post-training budget. Note that it is not a comparison of total computational cost or sample efficiency, because RL training and separate baseline tuning are excluded. Also, the RL score dispersion measures inference-seed variability conditional on the trained policies.

\begin{table}[htbp]
    \centering
    \caption{
    Best-found surrogate scores and baseline
    time-to-threshold statistics. Score means and standard deviations are calculated from the best
    score in each of three seeded runs, each with
    3,000 reward-model evaluations. RL runs use different
    inference seeds of one frozen trained policy per
    algorithm; baseline runs are separately initialized
    optimization runs. RL standard deviations measure inference-seed variability.
    Time to threshold is the first evaluation at which
    the running best reaches 95\% of that run's final
    within-budget best score, reported as mean $\pm$ SD
    across runs. No comparable time-to-threshold statistic
    is reported for pretrained RL policies. Pure Random
    is excluded because it has no surrogate evaluation.
    }
    \label{tab:convergence}
    \small
    \setlength{\tabcolsep}{6pt}
    \renewcommand{\arraystretch}{1.15}
    \begin{tabular*}{\textwidth}{@{\extracolsep{\fill}}l c c c@{}}
        \toprule
        \textbf{Method}
        & \textbf{Mean best score}
        & \textbf{SD}
        & \textbf{Evaluations to threshold} \\
        \midrule
        \multicolumn{4}{l}
        {\textit{Reinforcement-learning agents}} \\
        \midrule
        PPO           & 34.50 & 0.35 & -- \\
        RPPO          & 34.89 & 0.04 & -- \\
        TRPO          & 33.30 & 0.61 & -- \\
        \midrule
        \multicolumn{4}{l}
        {\textit{Optimization baselines}} \\
        \midrule
        L-SHADE       & 30.36 & 0.24 & $1648.7\pm514.2$ \\
        Random search & 27.52 & 0.54 & $57.0\pm53.7$ \\
        CMA-ES        & 26.18 & 1.63 & $878.3\pm952.4$ \\
        Ax BO         & 28.73 & 1.24 & $60.0\pm29.9$ \\
        RUN           & 29.87 & 0.77 & $2036.0\pm624.1$ \\
        PSO           & 29.29 & 0.60 & $491.3\pm275.4$ \\
        \bottomrule
    \end{tabular*}
\end{table}

\subsection{Enrichment of shortlists in high-performing
experimental neighborhoods}
\label{sec:neighborhood_results}

Higher surrogate scores do not necessarily indicate greater
experimental promise. We therefore assessed the top-ranked
candidate shortlists using measured devices from the
890-record neighborhood reference pool. A candidate was
covered if it had at least one reference neighbor satisfying
the similarity threshold. Covered candidates were classified
as Strong when their mean neighborhood PCE was at or above
the 85th percentile of the reference-pool PCE distribution
(approximately 22.86\%). Pure Random was included as a background reference
using 1,000 unselected samples, whereas the RL policies
and optimization baselines were evaluated using
100-candidate surrogate-ranked shortlists.

We summarize shortlist enrichment using the
neighborhood-supported hit rate,
$\mathrm{HR}_{\mathrm{support}}@K$:
the fraction of all $K$ shortlisted candidates that are
both covered and Strong. This metric incorporates both
reference coverage and support strength, rather than
conditioning only on covered candidates.

\begin{figure}[htbp]
    \centering
    \includegraphics[width=\textwidth]
    {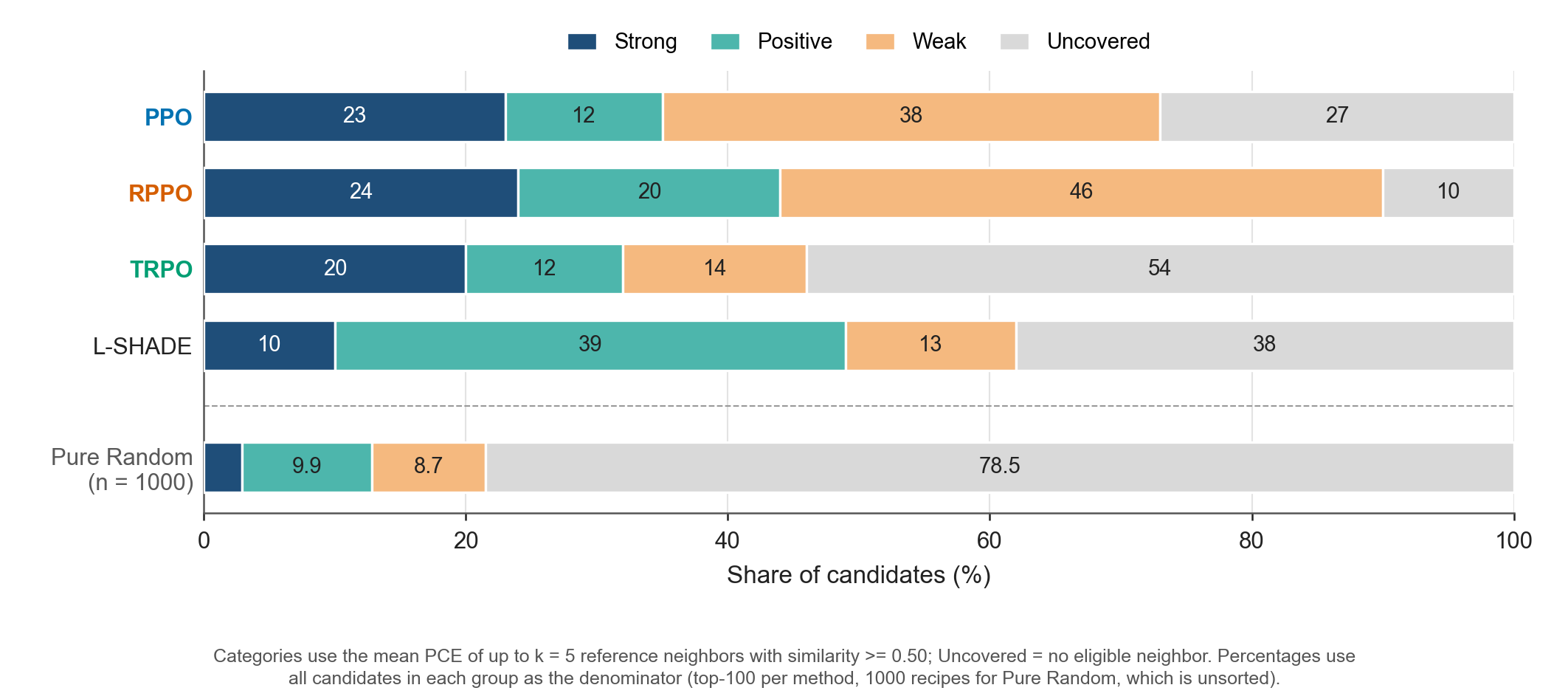}
    \caption{
    Reference-neighborhood evidence for the top-100
    surrogate-ranked candidates.
    Bars show the proportions classified as Strong, Positive, Weak,
    or Uncovered for PPO, RPPO, TRPO, and L-SHADE, the non-RL baseline
    with the highest best-found surrogate score
    (Table~\ref{tab:convergence}); L-SHADE is tied with Ax BO on the
    K = 100 supported hit rate (10\% each). All other baselines are
    shown in Figure~\ref{fig:neighbourhood_all_methods}.
    The lower bar shows Pure Random, a background reference of 1{,}000
    unselected recipes that were not surrogate-ranked; it is not a
    matched-shortlist comparison.
    Neighborhood PCE is the unweighted mean measured PCE of up to
    $k=5$ reference neighbors with similarity $\geq 0.50$.
    Strong denotes neighborhood PCE at or above the reference
    85th percentile; Positive denotes values at or above the
    reference median but below the 85th percentile; Weak denotes
    values below the median. Candidates with no eligible neighbor
    are Uncovered. Percentages use all candidates in each group as the
    denominator (100 per method; 1{,}000 for Pure Random).
    Segments below 7\% are not labeled; for Pure Random, Strong
    is 2.9\%.
    These categories describe reference-neighborhood evidence,
    not experimentally verified candidate performance.
    }
    \label{fig:neighbourhood}
    \label{fig:neighborhood_analysis}
\end{figure}

For the designated seed-42 top-100 shortlists at $k=5$, PPO and RPPO achieved
supported hit rates of 23\% and 24\%, respectively, while TRPO reached 20\%. Among
the non-RL optimizers, L-SHADE and Ax BO were tied at the highest rate (10\% each)
(Figure~\ref{fig:neighbourhood}). Thus, at $K=100$ and $k=5$, the supported hit rates of PPO and
RPPO were approximately 2.3-fold and 2.4-fold that of the non-RL optimizers with the highest supported hit rate (L-SHADE and Ax BO, 10\% each), corresponding to absolute differences of 13 and 14 percentage points. The corresponding neighborhood-support distributions
for all evaluated optimization methods are shown in
Figure~\ref{fig:neighbourhood_all_methods}.

The ordering persisted across the three evaluated maximum neighborhood sizes. PPO
and RPPO retained supported hit rates of 23\% and 24\% at $k=3$, 5, and 10. TRPO
reached 21\%, 20\%, and 20\%, compared with 10\%, 10\%, and 9\% for both L-SHADE
and Ax BO (Appendix~\ref{app:coverage-adjusted}). 
This sensitivity analysis supports the observed advantage
across the tested maximum neighbor counts; it does not
establish robustness to other similarity definitions,
thresholds, or reference pools.

These findings indicate that the learned policies generated shortlists with a larger fraction of high-PCE neighborhood-supported configurations than the evaluated optimization baselines. The results describe one designated candidate-generation or search seed per method.

\subsection{Early enrichment, coverage, and screening-quality trade-offs}
\label{sec:screening_tradeoffs}

A higher supported hit rate does not imply uniformly
better candidate quality. At $k=5$, coverage was 73\% for PPO, 90\% for RPPO, and 46\% for TRPO. Their
Uncovered fractions were therefore 27\%, 10\%, and 54\%, respectively.
Covered-and-Weak candidates accounted for 38\% of the complete PPO shortlist, 46\%
of the RPPO shortlist, and 14\% of the TRPO shortlist (Figure~\ref{fig:neighborhood_analysis}). Conditional on coverage, TRPO had the largest Strong fraction among the RL policies: 43.5\%, compared with
31.5\% for PPO and 26.7\% for RPPO. Its lower overall
coverage nevertheless reduced its supported hit rate.
Conversely, RPPO's broader coverage increased both the
number of Strong-supported candidates and the number
classified as Weak. Uncovered candidates lack an eligible
reference neighbor (i.e., fall outside the reference applicability domain) and are not assigned a low-performance
classification.

Mean neighborhood PCE among covered candidates provides
a complementary summary. At $k=5$, these means were $19.01\pm3.72\%$ for PPO, $18.70\pm3.54\%$ for RPPO, and
$20.33\pm3.61\%$ for TRPO, where the standard deviations describe variation among
candidate-level neighborhood means. L-SHADE and RUN reached $20.35\pm2.17\%$ and
$21.60\pm0.19\%$, respectively
(Appendix~\ref{app:coverage-adjusted}, Table~\ref{tab:signal_distribution}).

Thus, the policies with the highest supported hit rates did not have the highest
average neighborhood PCE among covered candidates. In particular, RUN had full
coverage and the highest mean neighborhood PCE, but 99 of its 100 candidates were
classified as Positive and only one as Strong. No RUN candidate had more than five
eligible neighbors (four had exactly five), so its $k=5$ and $k=10$ results are
identical. The methods
therefore differ in the distribution of reference
support, not simply in a single ordering of overall
screening quality. 

Enrichment relative to unselected sampling also partly
reflects reference coverage, because only 21.5\% of
Pure Random recipes were covered. Among covered candidates at $k=5$, the Strong fraction of Pure Random was 13.5\%,
compared with 43.5\% for TRPO, 31.5\% for PPO, and 26.7\% for RPPO, but 16.1\% for
L-SHADE and 17.5\% for Ax BO (Table~\ref{tab:signal_distribution}). The enrichment of L-SHADE and Ax BO
therefore arose largely from landing in covered regions, whereas PPO and RPPO
combined higher coverage (73\% and 90\%, versus 62\% for L-SHADE and 57\% for
Ax BO) with higher conditional support, and TRPO reached higher conditional support
despite lower coverage (46\%).

\begin{figure}[htbp]
    \centering
    \includegraphics[width=\textwidth]
    {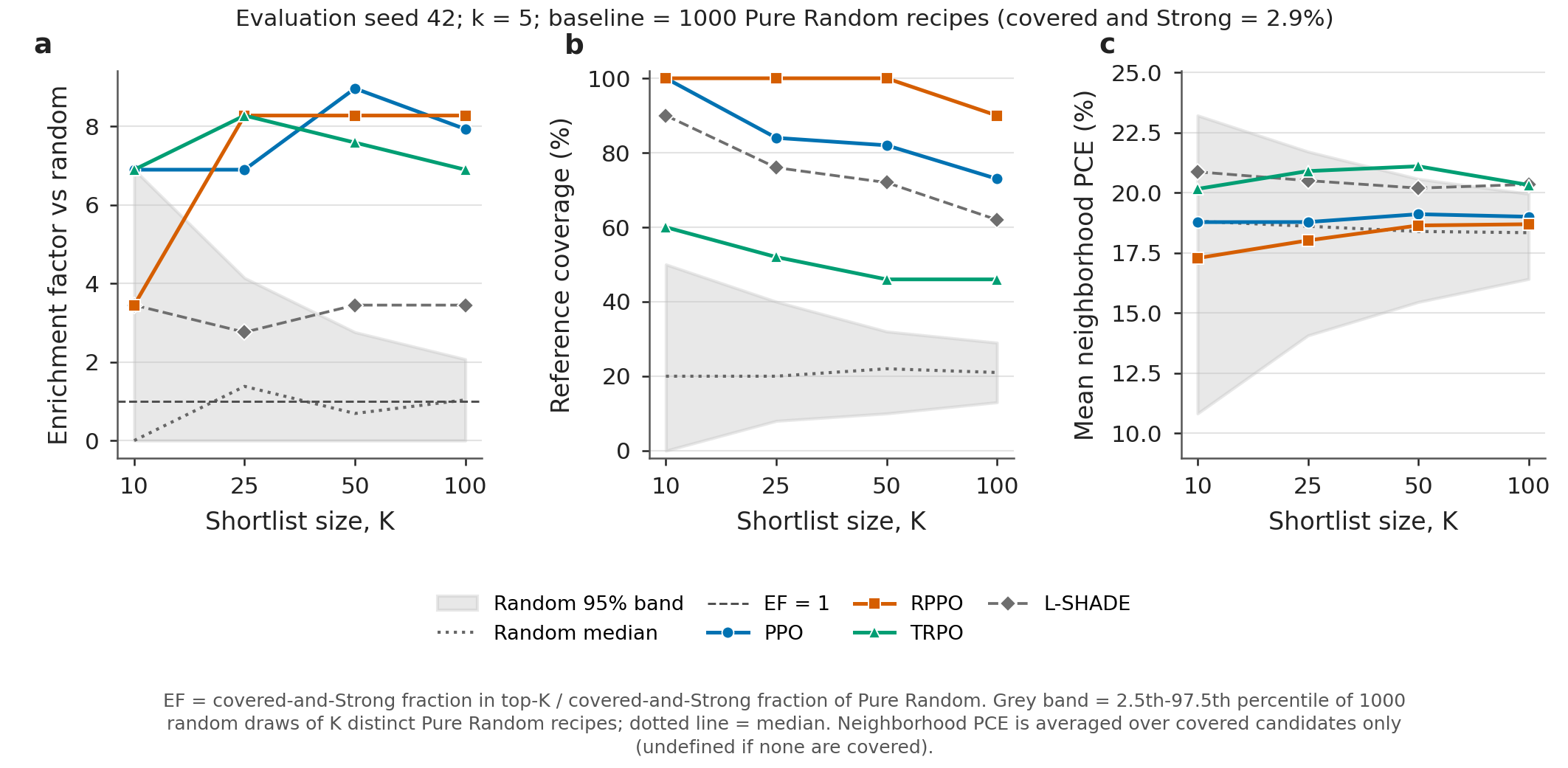}
    \caption{
    Early enrichment and screening-quality trade-offs.
    Candidates from PPO, RPPO, TRPO, and L-SHADE are ranked by
    descending saved surrogate-predicted PCE, and statistics are
    evaluated for top-K prefixes at K = 10, 25, 50, and 100.
    (a) Enrichment factor (EF): the neighborhood-supported hit
    rate (the fraction of all $K$ candidates that are covered
    and have mean neighborhood PCE at or above the
    reference-pool 85th percentile) divided by the same fraction
    among the 1{,}000 unselected Pure Random recipes (2.9\% at
    $k=5$). The dashed horizontal line marks $\mathrm{EF}=1$,
    the rate expected from unselected sampling.
    (b) Reference coverage: the fraction of all $K$ candidates
    with at least one eligible reference neighbor.
    (c) Mean candidate-level neighborhood PCE among covered
    candidates only; it is undefined when no candidate in the
    prefix is covered.
    In all panels, the grey band spans the 2.5th--97.5th
    percentiles of the statistic over 1{,}000 random draws of
    $K$ distinct recipes from the Pure Random sample, and the
    dotted line is the median of those draws. The band
    describes sampling variability of unselected selection
    only; it is conditional on the observed Pure Random sample.
    For each candidate, neighborhood PCE is the unweighted mean
    measured PCE of up to $k=5$ reference neighbors with
    similarity $\geq0.50$. $K$ denotes shortlist size and $k$
    the maximum number of reference neighbors.
    All results use the designated seed-42 run for each method.
    Lines connect evaluated shortlist sizes as visual guides
    only; they do not represent optimization trajectories or
    uncertainty across runs.
    These statistics describe experimental-reference support.
    }
    \label{fig:early_enrichment}
\end{figure}

To examine reference support at smaller shortlist sizes,
we evaluated neighborhood-supported hit rates, reference
coverage, and mean neighborhood PCE among covered candidates
for top-$K$ prefixes at $K=10$, 25, 50, and 100, using
a maximum neighborhood size of $k=5$. Hit rates were also
expressed as enrichment factors relative to unselected
Pure Random recipes, with the 95\% range of random
$K$-recipe selections shown for reference
(Figure~\ref{fig:early_enrichment}). The corresponding
all-method comparison is provided in
Figure~\ref{fig:early_enrichment_all_methods}.

No single policy had the highest supported hit rate at every shortlist size.
At $K=10$, 25, 50, and 100, the rates were 20\%, 20\%, 26\%, and 23\% for PPO;
10\%, 24\%, 24\%, and 24\% for RPPO; 20\%, 24\%, 22\%, and 20\% for TRPO; and
10\%, 8\%, 10\%, and 10\% for L-SHADE. The highest rate was shared by PPO and
TRPO at $K=10$, by RPPO and TRPO at $K=25$, belonged to PPO at $K=50$, and
belonged to RPPO at $K=100$. PPO and TRPO exceeded L-SHADE at every evaluated
shortlist size, whereas RPPO matched L-SHADE (and Ax BO) at 10\% for $K=10$ and
exceeded it at all larger sizes. The non-RL method with the highest supported hit rate was Ax BO at $K = 25$ and 50 (16\% and 14\%;
Figure~\ref{fig:early_enrichment_all_methods}); at $K=10$ and 100, L-SHADE and Ax BO were tied (10\%).
The ordering among RL policies changed with $K$: TRPO exceeded RPPO at $K=10$,
matched it at $K=25$, and fell below it at $K=50$ and 100. PPO's rate was lower
at $K=10$ and 25 (20\%) than at $K=50$ (26\%), and RPPO's top-ten rate (10\%) was
lower than its rates at larger shortlist sizes, indicating that its
Strong-supported candidates were not preferentially concentrated within the first
ten positions.

Relative to unselected sampling (2.9\% of Pure Random recipes covered and Strong
at $k=5$), the $K=100$ supported hit rates corresponded to enrichment factors of
7.9 for PPO, 8.3 for RPPO, 6.9 for TRPO, and 3.4 for L-SHADE (Figure~\ref{fig:early_enrichment}a). All
exceeded the upper limit of the random 95\% interval (2.1). At $K=50$ (limit 2.8),
all three RL policies (9.0, 8.3, and 7.6 for PPO, RPPO, and TRPO), L-SHADE (3.4),
and Ax BO (4.8) lay above the limit. At $K=25$ (limit 4.1), PPO (6.9), RPPO (8.3),
TRPO (8.3), and Ax BO (5.5) lay above it, whereas L-SHADE (2.8) lay within the
interval. At $K=10$ (limit 6.9), no method exceeded the limit: PPO and TRPO (both
6.9) lay at it, and RPPO, L-SHADE, and Ax BO (3.4 each) lay within it. The random
interval has a lower limit of zero at every $K$, so it indicates only whether a
method exceeds unselected sampling, not whether it falls below it.

Broader reference coverage did not necessarily translate into a higher supported
hit rate. At $K=50$, RPPO had 100\% coverage compared with 82\% for PPO, but its
supported hit rate was lower (24\% versus 26\%); at $K=10$, both had 100\%
coverage, but RPPO's rate was 10\% versus 20\% for PPO. At $K=25$, TRPO achieved
the same 24\% supported hit rate as RPPO with only 52\% coverage (versus 100\%).
Mean neighborhood PCE among covered candidates provided a different ordering. The
TRPO means were 20.17\%, 20.91\%, 21.11\%, and 20.33\%, and the L-SHADE means were
20.88\%, 20.51\%, 20.20\%, and 20.35\% at $K=10$, 25, 50, and 100. TRPO therefore
exceeded L-SHADE at $K=25$ and 50 but not at $K=10$ (where it rested on six
covered candidates) or $K=100$ (a difference of 0.02 percentage points). PPO and
RPPO had lower covered means than L-SHADE at every $K$, despite their higher
supported hit rates. Reference coverage also differed: at $K=25$, L-SHADE covered
76\% of candidates, compared with 84\% for PPO, 100\% for RPPO, and 52\% for TRPO.
Thus, early high-PCE support, reference coverage, and average neighborhood PCE
capture distinct aspects of screening quality.

\subsection{Composition-family agreement with separately
retrieved perovskite literature}
\label{sec:openalex-results}

\begin{figure}[htbp]
    \centering
    \includegraphics[width=\textwidth]
    {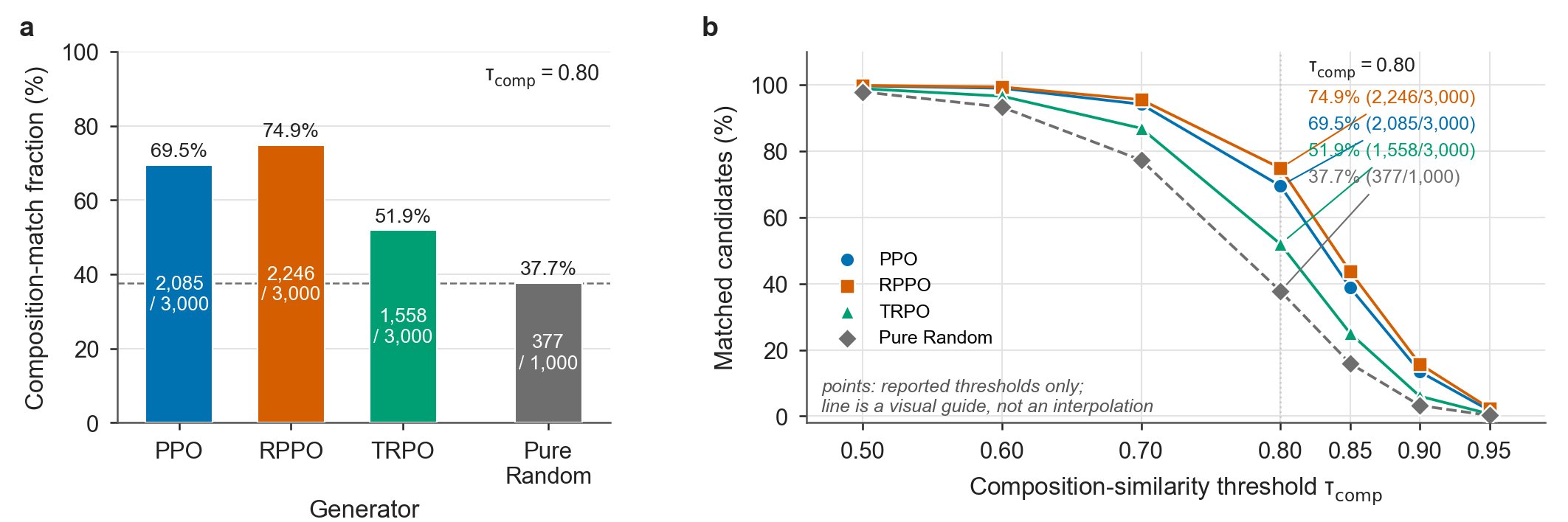}
    \caption{
    External composition-family matching against the OpenAlex
    reference pool.
    (a) Fractions of generated recipes whose best-match
    composition similarity meets the threshold
    \(\tau_{\mathrm{comp}} = 0.80\), shown separately for PPO,
    RPPO, and TRPO; labels report matched and total recipe
    counts.
    (b) Matching rates across the evaluated composition-similarity
    thresholds for the pooled 9,000 RL-generated recipes and
    1,000 unselected Pure Random recipes. At
    \(\tau_{\mathrm{comp}} = 0.80\), 5,889 RL-generated recipes
    (65.4\%) and 377 Pure Random recipes (37.7\%) meet the
    threshold. At 0.70, the corresponding rates are 92.2\%
    and 77.3\%; at 0.90, they are 11.7\% and 3.2\%.
    Connecting lines are visual guides only. Matching concerns
    composition-family similarity only. Detailed numerical results are
    provided in Appendix~\ref{app:openalex}.
    }
    \label{fig:composition_matching}
\end{figure}

We characterized the absorber compositions of 9,000 generated
recipes using a composition reference pool of \(n = 601\)
retrieved through OpenAlex. This analysis used the full
seed-42 RL candidate pools rather than the top-100 shortlists (Section~\ref{sec:baselines_metrics}).
At the operational similarity threshold $\tau_{\mathrm{comp}} = 0.80$,
5,889 recipes (65.4\%) matched a retrieved composition family
(RPPO: 74.9\%, PPO: 69.5\%, TRPO: 51.9\%; Fig.~\ref{fig:composition_matching}a).
Applying the same criterion to 1,000 unselected Pure Random recipes
yielded 377 matches (37.7\%). The pooled generated-recipe matching rate was 27.7 percentage points higher than the unselected background rate.

Agreement was strongly threshold-dependent. For the pooled RL-generated recipes, the matching rate was 92.2\% at \(\tau_{\mathrm{comp}} = 0.70\), 65.4\% at 0.80, 35.8\% at 0.85, 11.7\% at 0.90, and 1.6\% at 0.95. The corresponding Pure Random rates were 77.3\%, 37.7\%, 15.9\%, 3.2\%, and
0.2\%, respectively. These results show greater agreement with the retrieved composition reference pool for the
pooled RL-generated recipes than for unselected sampling
across the evaluated thresholds, while also demonstrating that the absolute matching rate depends strongly on the chosen cutoff.

These results characterize the compositional regions occupied by the proposals. It is possible that recent papers adopt composition from previous literature. Thus, restricting
retrieval to recent non-open-access works reduces
intended source overlap but does not prove independence from the training data at the composition level.

\section{Discussion}
\label{sec:discussion}

\subsection{From surrogate optimization to experimentally informed screening}
\label{subsec:disc_summary}

This study asked whether policies trained on a frozen surrogate can serve as the
candidate-generation stage of a virtual-screening workflow, and whether the
resulting shortlists are enriched in configurations supported by high-PCE
experimental neighborhoods. Three findings address this question.

\begin{enumerate}
  \item \textbf{Both screening signals carry ranking information on measured
  devices.} Top-ten selection raised the mean measured PCE from 19.68\% to
  23.74\%, and neighborhood-mean PCE correlated with measured PCE at Spearman
  $\rho \approx 0.68$--$0.71$.
  \item \textbf{All three policies reached higher best-found surrogate scores
  (33.30--34.89) than any baseline ($\leq 30.36$)} within 3{,}000 reward-model
  evaluations.
  \item \textbf{The policies enriched shortlists in neighborhood-supported
  configurations.} At $K = 100$ and $k = 5$, PPO and RPPO reached supported hit
  rates of 23\% and 24\%. These are approximately 2.3- and 2.4-fold that of the
  strongest non-RL optimizers (L-SHADE and Ax BO, tied at 10\%), and 7.9- and
  8.3-fold that of unselected sampling (2.9\%).
\end{enumerate}

The neighborhood criterion was never part of the reward. The enrichment
therefore emerges from surrogate optimization alone, which suggests that the
surrogate and the experimental neighborhoods point toward overlapping regions of
the design space. The following subsections discuss why this might occur, how
the policies differ, and how the framework can be used. The boundaries of the
evidence are collected in Section~\ref{subsec:limitations}.

\subsection{Factors that may contribute to the enrichment}
\label{subsec:disc_origins}

\paragraph{Decision formulation.}
Each action specifies a complete encoded recipe and $\gamma = 0$. Maximizing the
objective is therefore equivalent to maximizing the surrogate score of the next
recipe (Section~\ref{sec:methods}). The trained policies act as proposal
distributions over high-scoring recipes. This suits a screening workflow, which
needs a reusable generator of good candidates more than a sequential planner.
The decrease in policy entropy during training (entropy loss from $-71.8$ to
$-55.1$ for PPO and to $-51.8$ for RPPO) is consistent with this view. The
discount-factor ablation points the same way. Rewards stayed near the
random-policy level at $\gamma = 0.99$, whereas $\gamma = 0$ produced
substantial improvement in all three algorithms
(Appendix~\ref{app:gamma}).

\paragraph{Transfer of literature knowledge through the surrogate.}
The TabPFN reward model conditions on 1{,}292 literature records, so regions it
scores highly partly coincide with regions populated by high-PCE literature
devices. These are also the regions where a literature-derived reference pool is
most likely to contain neighbors. This overlap is a plausible mechanism by which
optimizing the surrogate yields neighborhood-supported shortlists. The baselines
show part of the same pattern. L-SHADE and Ax BO gained mainly from landing in
covered regions, whereas PPO and RPPO combined higher coverage (73\% and 90\%,
versus 62\% and 57\%) with higher conditional support.

\paragraph{Pre-trained, reusable generators.}
The policies are trained once and then produce candidates at inference time.
Freshly initialized optimizers must search from scratch on each run. In this
setting, the learned policies generate high-scoring, neighborhood-supported
candidates at inference time without a new search.

\subsection{Policies offer complementary support profiles}
\label{subsec:disc_quality}

The supported hit rate is the product of coverage and conditional support, and
the three policies reach it in different ways.

\begin{itemize}
  \item \textbf{RPPO} had the broadest coverage (90\%) and the highest supported
  hit rate at $K = 100$ (24\%). It is a good default when the aim is to maximize
  the number of Strong-supported candidates.
  \item \textbf{TRPO} had the highest conditional Strong fraction (43.5\%) and
  the lowest Weak share (14\% of its shortlist). It suits users who want the
  most reliable support among covered candidates, and it also leaves more
  candidates in unexplored regions.
  \item \textbf{PPO} lay between the two (73\% coverage, 31.5\% conditional
  Strong). It had the highest rate at $K = 50$ (26\%) and tied for the highest
  at $K = 10$.
\end{itemize}

These profiles differ from those of conventional optimizers. RUN and L-SHADE
reached high mean neighborhood PCE among covered candidates (21.60\% and
20.35\%), concentrated in the Positive band. The learned policies instead put
more candidates in the top support tier. Mean neighborhood PCE, coverage, and
early high-PCE support capture distinct aspects of screening quality. The
supported hit rate is the one most directly tied to prioritizing candidates for
high performance.

From $K = 25$ upward, PPO, RPPO, and TRPO all exceeded the upper limit of the
random 95\% interval at every evaluated shortlist size. Their enrichment factors
ranged from 6.9 to 9.0 at $K = 50$ and $K = 100$.

\subsection{Surrogate and neighborhood scores as complementary signals}
\label{subsec:disc_exploitation}

The surrogate and the neighborhood score answer different questions. The
surrogate is a smooth model that can be queried anywhere in the encoded space.
The neighborhood score compares a candidate with measured devices, but only
where measured analogues exist. Within-fold ranking by the surrogate (Spearman
0.436) and the internal neighborhood--PCE association ($\rho \approx 0.70$) both
showed useful ranking information. Surrogate rank did not order reference
support within a shortlist. For example, RPPO's top-ten supported hit rate
(10\%) was lower than at larger $K$. This is consistent with the two signals
being distinct, which suggests that combining them could add information beyond
either alone.

A candidate with both a high surrogate score and Strong support is therefore the
most credible type of proposal. A candidate with a high surrogate score and no
eligible neighbor (Uncovered) is a natural exploratory slot. Best-found
surrogate scores are used to rank methods under a common reward and are not
interpreted as device efficiencies.

\subsection{Composition-family agreement}
\label{subsec:disc_composition}

At $\tau_{\mathrm{comp}} = 0.80$, 65.4\% of RL-generated recipes matched a
composition family retrieved from recent non-open-access literature, compared
with 37.7\% of unselected Pure Random recipes. The ordering
RPPO $>$ PPO $>$ TRPO, and the excess over Pure Random, held at every evaluated
threshold from 0.50 to 0.95. The policies therefore concentrate in
compositional regions that recent studies also occupy, instead of scattering
across the ratio space. This supports the plausibility of the proposals at the
level of absorber composition.

The framework is an exploitation-oriented screen. It recombines known
compositions, layers, and processes to find configurations where surrogate and
experimental neighbors agree, which is the intended role of a screening stage.
Claims about novelty are addressed in Section~\ref{subsec:limitations}.

\subsection{Implications for experimental follow-up}
\label{subsec:disc_followup}

The immediate use of the framework is to prioritize candidates for experimental
follow-up. A practical batch would combine
\begin{enumerate}
  \item a high surrogate score;
  \item strong reference support;
  \item diversity in composition and layer configuration; and
  \item a few deliberately exploratory slots filled by high-scoring Uncovered
  candidates.
\end{enumerate}
Fabricating such a batch alongside one selected by a baseline optimizer under
the same budget would directly test whether reference-supported prioritization
improves screening outcomes, and whether the surrogate and support signals
predict real devices. The primary endpoint should be defined in advance, and all
devices should be measured under standard illumination.

The pipeline is modular. The reference pool is used only after generation, so it
can be extended with new measurements and the support analysis rerun without
retraining the policies. Additional records would also raise the coverage of
unselected recipes (currently 21.5\%) and widen the region where support can be
assessed. A natural extension is to rank by both surrogate score and reference
support. This would require a held-out reference pool, so that support remains
an independent assessment.

\subsection{Limitations}
\label{subsec:limitations}

\paragraph{Evidence type and surrogate behavior.}
\begin{itemize}
  \item \textbf{Computational evidence only.} Strong, Positive, Weak, and
  Uncovered describe the measured PCE of neighboring reference devices, not of
  the proposed devices. Weak does not establish experimental failure, and
  Uncovered indicates only that no eligible neighbor was found. Because support
  is defined by similarity to known devices, a high Strong fraction can result
  from proposing recipes close to existing high-PCE records. Reference support
  therefore does not establish novelty.

  \item \textbf{Surrogate accuracy and exploitation.} The surrogate has moderate
  accuracy ($R^2 = 0.414$). In cross-validation it underpredicts known high
  performers (mean bias $-0.87$ pp for 20--22\% devices and $-2.55$ pp above
  24\%, with only 35 records above 24\%). This is consistent with regression
  toward the training mean and gives no assurance about out-of-distribution
  behavior. The optimizers reach scores of 30--35, well beyond the measured
  range. Even random search reached 27.52, above the highest PCE in the
  reference pool (26.83\%). The best-found scores of the RL policies
  (33.30--34.89) reach or exceed the single-junction Shockley--Queisser limit
  (approximately 33\%), so they likely reflect some surrogate exploitation. They
  are informative only for ranking methods under a common reward, not as
  predicted device efficiencies.
\end{itemize}

\paragraph{Statistical scope and comparison design.}
\begin{itemize}
  \item \textbf{One run per method.} The enrichment results come from one
  designated shortlist per method, without independent policy-training runs.
  The study therefore does not establish seed robustness. The reported
  dispersion of RL scores measures inference-seed variability conditional on the
  trained policies. The ordering among RL policies changed with $K$, and the
  composition-matching ordering (RPPO $>$ PPO $>$ TRPO) likewise rests on a
  single seed-42 pool per policy.

  \item \textbf{Unmatched pre-training budgets.} Policy training used 819{,}200 surrogate
  evaluations, whereas each baseline run used 3{,}000. Baseline tuning was
  limited (eight Optuna trials for three baselines and none for the other
  three), and neither RL training nor baseline tuning is counted in the
  evaluation budget. The comparison is between pre-trained, reusable generators
  and freshly initialized optimizers. It does not show that the learned policies
  are more sample-efficient or cheaper, and part of the observed difference may
  reflect this asymmetry.

  \item \textbf{Decision formulation.} With $\gamma = 0$ and full-recipe
  actions, the policies act as proposal distributions, not as long-horizon
  planners. The results do not show that planning or recurrence is beneficial,
  because PPO and RPPO performed similarly and their ordering changed with
  shortlist size. The discount-factor ablation reflects one training run per
  setting, and the proposed mechanism for the $\gamma = 0.99$ failure
  (Appendix~\ref{app:gamma}) is illustrative only.
\end{itemize}

\paragraph{Interpretation of enrichment.}
\begin{itemize}
  \item \textbf{Non-independence of surrogate, reference pool, and similarity
  weights.} The reference pool and the reward-model data derive from related
  literature, and the reference pool excludes records eligible under the
  reward-model complete-case rule (Section~\ref{sec:dataset}) but is drawn from
  related sources. The similarity weights come from a Random Forest fitted on
  the same 1{,}292 records as the reward model. The neighborhood analysis is
  therefore a complement to the surrogate and not fully independent evidence. We
  cannot determine how much of the enrichment reflects overlap between regions
  the surrogate favors and regions where the reference pool has neighbors. The
  behavior of the baselines is compatible with this only in part. The enrichment
  of L-SHADE and Ax BO arose largely from landing in covered regions
  (conditional Strong fractions of 16.1\% and 17.5\%, against 13.5\% for Pure
  Random), whereas the PSO and RUN shortlists concentrated on a few reference
  devices.

  \item \textbf{Enrichment partly reflects coverage.} Only 21.5\% of Pure Random
  recipes were covered, and the supported hit rate counts Uncovered candidates
  as non-Strong. The learned policies' advantage is specific to this
  high-support criterion and is not a general superiority across quality
  measures. RUN (21.60\%) and L-SHADE (20.35\%) had mean neighborhood PCE among
  covered candidates at or above TRPO (20.33\%) and above PPO and RPPO (19.01\%
  and 18.70\%). RUN's value reflects 99 of 100 candidates lying in the Positive
  band and not a strong upper tail, and its shortlist occupied a narrow
  composition region.

  \item \textbf{Substantial Weak fractions.} Weak candidates made up 38\% of the
  PPO shortlist and 46\% of the RPPO shortlist. Surrogate optimization alone
  therefore also steers candidates toward regions where the nearest measured
  devices perform modestly. Our analysis cannot distinguish whether these cases
  are surrogate artifacts or genuinely new configurations that the literature
  does not yet support.

  \item \textbf{Weak evidence at small shortlist sizes.} At $K = 10$, no method
  exceeded the upper limit of the random 95\% interval, and PPO and TRPO only
  reached it. This is close to the batch size of many fabrication campaigns, so
  the enrichment is least demonstrated where it would be most useful, and it
  becomes clearer from $K = 25$ upward. Surrogate rank also did not order
  reference support within a shortlist. RPPO's top-ten supported hit rate (10\%)
  was lower than at larger $K$, so its highest-scoring candidates were not
  preferentially its best-supported ones.

  \item \textbf{Strong labels show resemblance only.} A Strong label shows
  resemblance to high-PCE devices and does not establish that the candidate will
  perform well. The Strong, Positive, and Weak thresholds are quantiles of the
  reference pool ($Q_{0.85}$ and $Q_{0.50}$), so they are relative to this
  particular pool.
\end{itemize}

\paragraph{Composition-family matching.}
\begin{itemize}
  \item \textbf{Weak null and post hoc threshold.} Pure Random samples uniformly
  over discrete ratios and is a weak null. The 27.7 pp contrast at
  $\tau_{\mathrm{comp}} = 0.80$ is the largest across the evaluated thresholds,
  and $\tau_{\mathrm{comp}}$ was selected after inspecting the results. It
  represents the upper end of the contrast and not a pre-specified test. The
  threshold-independent finding is the ordering of the policies and their excess
  over Pure Random at every threshold.

  \item \textbf{Dependence on the training data.} Recent papers may adopt
  compositions from earlier literature. Restricting retrieval to recent
  non-open-access works reduces intended source overlap but does not prove
  independence from the training data.

  \item \textbf{Strong threshold dependence.} Agreement fell steeply with
  threshold (11.7\% at 0.90 and 1.6\% at 0.95), so the absolute matching rate
  depends strongly on the cutoff.

  \item \textbf{Familiarity, not novelty.} Most proposals are close relatives of
  known families, as expected from a seven-species vocabulary. The results are
  not evidence of discovery of new chemistry. Matching relies on information
  extractable from titles and abstracts. It cannot establish literature-wide
  novelty or complete-device feasibility, and a familiar composition may be
  paired with a poorly supported layer configuration. A family match also does
  not rule out surrogate exploitation.
\end{itemize}

\paragraph{Data and representation.}
\begin{itemize}
  \item \textbf{Experimental records.} Records are affected by missing
  information, inconsistent reporting, and publication bias. Retaining the
  highest reported PCE per recipe defines a target that does not represent
  average device performance. The heuristic tandem screen was applied to the
  reference pool but not to the reward-model dataset. The light-intensity
  restriction was applied only to Perovskite Database records, so it covers 126
  of the 890 reference records. The other 764 were LLM-extracted and some may
  have been measured under non-standard illumination. The neighborhood
  assessment therefore relies mainly on the extraction pipeline.

  \item \textbf{Band-gap imputation.} Band gaps were missing for 39.1\% of the
  high-PCE records and were imputed from composition. The imputation model's
  80:20 split (Appendix~\ref{app:bandgap}) was drawn independently of the 70:30
  reward-model split, so devices held out for PCE evaluation may have
  contributed to its training set. The effect of this overlap on held-out PCE
  error cannot be isolated.

  \item \textbf{Representation.} Precursor chemistry, solvents, annealing
  conditions, and layer thicknesses are not encoded. The seven-species
  composition vocabulary excludes several minority cations, spacer cations, and
  halide or pseudohalide species, and the \texttt{Other} category obscures
  specific materials. Stoichiometric normalization is an encoding constraint. It
  does not establish phase stability, synthesizability, or layer compatibility.

  \item \textbf{Similarity metric and reference coverage.} Similarity weights
  come from an auxiliary Random Forest and not from the TabPFN reward model. The
  association between neighborhood scores and measured PCE was established only
  within the reference pool (coverage 96.4\%) and does not directly validate PCE
  predictions for generated devices. Coverage was much lower for unselected
  samples (21.5\%), and missing textual descriptors can lower similarity
  independently of candidate quality. We therefore report both the Strong
  fraction among covered recipes and the supported hit rate. The B-site elements
  (Pb and Sn) receive a combined weight below 0.07 of a total of 1.04.
  Sn-containing candidates, including the RUN shortlist (Cs $\approx 0.05$,
  FA $\approx 0.58$, MA $\approx 0.36$, Sn-only B-site), can therefore match
  neighbors that contain no Sn. We did not verify the B-site chemistry of the
  neighbors, so support ratings for these candidates should be interpreted
  cautiously.

  \item \textbf{Diversity of proposals and evidence.} Uniqueness after
  normalization does not imply diversity. All 88 covered PSO candidates shared
  one reference neighbor (neighborhood PCE 17.90\%), and the RUN neighborhoods
  were drawn from a few reference devices.
\end{itemize}

\subsection{Concluding remarks}
\label{subsec:disc_conclusion}

Surrogate-trained reinforcement-learning policies produced shortlists with a
substantially larger share of high-PCE, neighborhood-supported configurations
than the six optimizers tested. At $K = 100$ the supported hit rates were
23--24\% for PPO and RPPO, against 10\% for the best baselines and 2.9\% for
unselected sampling. Because the neighborhood criterion was never part of the
reward, this enrichment is a useful sign that surrogate optimization points
toward experimentally supported regions of the design space. The framework
offers a practical, modular way to prioritize candidates for fabrication.
Whether it improves measured device performance remains to be tested, as
discussed in Section~\ref{subsec:limitations}.

\section{Methods}
\label{sec:methods}

\subsection{Experimental data and recipe representation}
\label{sec:dataset}
\label{sec:recipe_formulation}

\paragraph{Data sources and extraction.}
The Perovskite Database~\cite{jacobsson2022perovskite}
contains 43,398 device records describing absorber compositions,
device architectures, fabrication procedures, and photovoltaic
performance. Its records were drawn primarily from papers published
before 2020.

We supplemented the database with recipe information extracted from
recent open-access papers using a large language model (LLM).
Extraction used a schema-constrained JSON prompt followed by
programmatic validation. After validation, eligibility filtering,
and deduplication, 667 LLM-extracted recipes were included in the
1,846-record high-PCE dataset used for reward-model development and
evaluation. The extraction prompt, output schema, and validation
procedure are provided in Appendix~\ref{app:extraction}.

\paragraph{Reward-model datasets.}
Eligible records had a reported PCE, a composition that passed the
encoding and stoichiometry checks, and complete values for all
selected features except band gap. Records from the Perovskite
Database were restricted to reported light intensities of
90--110~mW\,cm$^{-2}$. The LLM-extracted records were not filtered by
light intensity; approximately 40\% of the raw extracted records
reported this descriptor (Appendix~\ref{app:extraction}).

We constructed a full cleaned dataset without a lower PCE threshold
or publication-year restriction and a high-PCE dataset restricted
to PCE $\geq 18\%$ and publication year $\geq 2016$.
Layer-material strings were harmonized before encoding and
deduplication. For example, Spiro-OMeTAD was mapped to Spiro-MeOTAD.
Duplicate recipes were identified by composition, deposition
procedures, and layer-stack sequences. For each key, we retained
the record with the highest reported PCE, making the maximum
reported PCE the prediction target for the encoded recipe.
For the high-PCE dataset, recipe-level deduplication preceded the
publication-year filter. The resulting dataset contained
1,846 records. Additional cleaning and feature-preparation details
are provided in Appendix~\ref{app:data}.

\paragraph{Experimental-neighborhood reference pool.}
We assembled a separate reference pool from Perovskite Database
and LLM-extracted records for neighborhood assessment of generated
candidates. Eligible records had a reported PCE, a publication
year $\geq 2020$, and a composition that passed the encoding and
stoichiometry checks. No lower PCE threshold was imposed.
Perovskite Database records were restricted to reported light
intensities of 90--110~mW\,cm$^{-2}$, while LLM-extracted records
were not filtered by light intensity.

Reference-pool records were required to have at least one missing
value across eight descriptors: the four layer-stack sequences
(substrate, ETL, HTL, and back contact) and the four deposition
procedures (perovskite, ETL, HTL, and back contact). This
missing-descriptor criterion excluded records eligible under the
reward-model complete-case rule.

As data-quality screens, we excluded records with reported PCE
above 27\% and records with stack strings indicating possible
tandem devices. These screens were applied only to the reference
pool.

We deduplicated the reference pool using the same key as for the
reward-model datasets: perovskite composition, the four
layer-stack sequences, and the four deposition procedures. For
each key, we retained the record with the highest reported PCE.
Missing band gaps were imputed from composition. The final pool
contained 890 measured-device records: 764 LLM-extracted records
and 126 Perovskite Database records. Further construction details
are provided in Appendix~\ref{app:reference-pool}, and
component-specific missing-value handling is described in
Appendix~\ref{app:similarity}.

The OpenAlex composition reference pool used for literature
matching is described in Section~\ref{sec:openalex-method}.

\paragraph{Recipe representation.}
We selected 12 database features using domain knowledge and fixed
the feature set before reward-model training without using PCE
targets for feature selection. The features comprised 10
agent-controlled base features, the perovskite band gap, and the
publication year. The agent-controlled features expanded into a
21-dimensional action and observation vector: 13 categorical
layer-material and deposition entries, seven composition entries,
and one quenching flag. The feature list and vector layout are
provided in Tables~\ref{tab:features} and~\ref{tab:vector_layout}.
The agent operated in a discrete \texttt{MultiDiscrete} action
space. Categorical recipe entries were target-encoded for
reward-model evaluation using mappings fitted on the corresponding
training partition. Detailed encoding rules are provided in
Appendix~\ref{app:data:vector}.

\begin{table}[htbp]
    \centering
    \caption{Selected base features and their representation.
    Ten agent-controlled features expand into the
    21-dimensional recipe vector. Band gap and publication year
    are additional reward-model inputs.}
    \label{tab:features}
    \small
    \setlength{\tabcolsep}{6pt}
    \renewcommand{\arraystretch}{1.15}
    \begin{tabular*}{\textwidth}{@{\extracolsep{\fill}}l l l@{}}
        \toprule
        \textbf{Feature} & \textbf{Type} & \textbf{Encoding} \\
        \midrule
        Perovskite composition      & Composition & 7-dim ratio vector \\
        Perovskite band gap         & Numerical   & Not transformed \\
        Substrate stack sequence    & Sequence    & Positional, \(\le 2\) layers \\
        ETL stack sequence          & Sequence    & Positional, \(\le 3\) layers \\
        HTL stack sequence          & Sequence    & Positional, \(\le 2\) layers \\
        Back-contact stack sequence & Sequence    & Positional, \(\le 2\) layers \\
        Perovskite deposition       & Categorical & Target-encoded \\
        ETL deposition              & Categorical & Target-encoded \\
        HTL deposition              & Categorical & Target-encoded \\
        Back-contact deposition     & Categorical & Target-encoded \\
        Quenching-induced crystallization & Boolean & \(0 / 1\) \\
        Publication year            & Integer     & Not transformed \\
        \bottomrule
    \end{tabular*}
\end{table}

Layer sequences were split into positional sublayers. We retained
up to three ETL sublayers and up to two sublayers each for the
substrate, HTL, and back contact. Rare layer materials were grouped
into an \texttt{Other} category.

Composition was represented by
$(\mathrm{Cs},\mathrm{FA},\mathrm{MA},
\mathrm{Pb},\mathrm{Sn},\mathrm{Br},\mathrm{I})$.
Raw discrete composition entries were normalized to satisfy the
represented $\mathrm{ABX_3}$ stoichiometry:
\[
\mathrm{Cs}+\mathrm{FA}+\mathrm{MA}=1,\qquad
\mathrm{Pb}+\mathrm{Sn}=1,\qquad
\mathrm{Br}+\mathrm{I}=3.
\]

\subsection{Surrogate reward and its validation}
\label{sec:reward_model}

\paragraph{Model inputs and configuration.}
We used \texttt{TabPFNRegressor}~\cite{hollmann2025tabpfn}
to predict PCE. TabPFN performs in-context prediction using a
fitted training context. We used its default configuration
without hyperparameter tuning.

The reward model used 23 inputs: the 21 recipe entries,
perovskite band gap, and publication year. Categorical
preprocessing comprised rare-category handling, ordinal encoding,
target encoding, and a quantile transformation, as detailed in
Appendices~\ref{app:data:vector} and~\ref{app:reward_model}.

\paragraph{Band-gap imputation.}
A separate TabPFN regressor predicted band gap from the seven
composition features (Cs, FA, MA, Pb, Sn, Br, and I).
The standalone band-gap model described in
Appendix~\ref{app:bandgap} achieved a held-out MAE of 0.015~eV
and $R^2=0.979$.

Of the 1,846 high-PCE records, 722 (39.1\%) lacked a reported
band gap. Within each PCE cross-validation fold, we fitted a
fresh band-gap model using training-fold records with observed
band gaps. The model imputed missing band gaps in both the
training and validation partitions. Observed band gaps were
retained unchanged.

\paragraph{Cross-validation and ranking assessment.}
We evaluated predictive performance separately on the full
cleaned dataset and the high-PCE dataset using five-fold
stratified cross-validation. For the high-PCE dataset,
stratification used 20 quantile bins of measured PCE, with
shuffling and random seed 50. The validation folds contained
370, 369, 369, 369, and 369 records.

In each fold, preprocessing operations were fitted on the
training partition and applied to the validation partition.
The same 23 input features were used in every fold. Each
eligible record received exactly one out-of-fold prediction.

We reported the mean and standard deviation across folds for
$R^2$, MAE, and RMSE, together with the corresponding metrics
calculated from pooled out-of-fold predictions. Within each
validation fold, we assessed ranking using Spearman correlation
and the mean measured PCE of the ten highest-scoring records.
When scores were tied at the selection cutoff, the top-ten
statistic used the expected mean under uniform selection among
the tied records competing for the remaining positions.

The random-selection reference was the mean measured PCE of
the validation fold, equal to the expected mean of ten records
sampled uniformly without replacement. Selection gain was
defined as the difference between the top-ten mean and this
reference, expressed in percentage points of PCE.

\paragraph{Frozen surrogate used for optimization.}
For recipe optimization, we split the 1,846-record high-PCE
dataset into 1,292 training records and 554 held-out records
using a stratified $70{:}30$ split with random seed 50 and the
same 20 PCE quantile bins. Before this split, missing band gaps
were filled using the saved band-gap surrogate described in
Appendix~\ref{app:bandgap}. This surrogate used the seven
composition features as inputs.

The ordinal-encoding vocabulary and the
singleton-to-\texttt{Other} mapping were derived from all
1,846 records before splitting, using only category identities
and frequencies. The target encoder, quantile transformer, and
PCE model were fitted on the 1,292 training records. The frozen
model achieved $\mathrm{RMSE}=1.156$ percentage points on the
554 held-out records in this single-split evaluation.

The PCE model, preprocessing transforms, and saved band-gap
surrogate were held fixed during optimization. For generated
candidates, band gap was derived from composition and publication
year was set to 2026. This pipeline defined the surrogate score
$R(s)$ for an encoded recipe $s$.

\subsection{RL-based recipe generation}
\label{sec:rl_agent}

\paragraph{Decision formulation.}
The recipe environment was implemented using
Gymnasium~\cite{towers2024gymnasium}. Each action specified the
complete next encoded recipe rather than a local modification.
The environment decoded the action, normalized its composition,
augmented the recipe with band gap and publication year, and
evaluated the frozen surrogate.

Using recipe-level notation $s_{t+1}=a_t$, the reward was
\begin{equation}
    r_t=R(s_{t+1})-R(s_t),
    \label{eq:recipe_reward}
\end{equation}
where $R(s)$ is the surrogate-predicted PCE for the augmented
recipe at evaluated states.

The main experiments used $\gamma=0$. The objective at state $s$
was therefore
\[
\mathbb{E}_{a\sim\pi(\cdot\mid s)}
\left[R(a)-R(s)\right].
\]
Because $R(s)$ is constant with respect to the action at that
state, maximizing this objective is equivalent to maximizing
the expected surrogate score of the next recipe.

Each episode started from the all-zero reset state $s_0$.
We assigned this state a fixed reference value $R(s_0)=35$
without querying the surrogate. The first reward was
$r_0=R(s_1)-35$; subsequent rewards were computed from the
surrogate scores of successive recipes. The reset state was
excluded from the candidate pools.

The undiscounted episode reward used in
Table~\ref{tab:gamma_ablation} was
\[
\sum_{t=0}^{T-1}r_t=R(s_T)-35,
\]
where $s_T$ is the final recipe in the episode. The policy updates
in the main experiments used the $\gamma=0$ objective above.

\paragraph{Policies and training.}
We trained PPO~\cite{schulman2017ppo}, recurrent PPO (RPPO), and
TRPO~\cite{schulman2015trpo} using
Stable-Baselines3~\cite{stable-baselines3} and
\texttt{sb3-contrib}~\cite{sb3-contrib}. PPO and TRPO used
multilayer-perceptron policies; RPPO additionally used a
long short-term memory (LSTM) network.

Each training run requested $8\times10^5$ environment steps,
using 100 parallel environments and 256 steps per environment
per rollout. Completing the final rollout gave 819,200 actual
environment steps. Episodes ended after 200 steps or when the
predicted-PCE change satisfied the $10^{-4}$ stopping criterion
for 15 consecutive steps. The differential reward was the sole
reward term.

All implementations used generalized advantage
estimation (GAE)~\cite{schulman2016gae} with $\lambda=0.98$.
At $\gamma=0$, the advantage estimate reduces to the one-step
temporal-difference residual. We also evaluated $\gamma=0.99$
under otherwise identical settings.

Training hyperparameters are summarized in
Table~\ref{tab:rl_hyperparams}. Composition-normalization details,
including the zero-sum fallback, are provided in
Appendix~\ref{app:hyperparams}. Training diagnostics and the
discount-factor comparison are reported in
Appendices~\ref{app:training_diagnostics} and~\ref{app:gamma},
respectively.

\paragraph{Post-training generation.}
Policy parameters were frozen during candidate generation.
The surrogate score defined the reward; neighborhood support
and composition-family matching were evaluated after generation.
Candidate-scoring budgets, shortlist construction, and evaluation
seeds are specified in Section~\ref{sec:baselines_metrics}.

\begin{table}[htbp]
    \centering
    \caption{Reinforcement learning hyperparameters. All three
    algorithms use the same environment, reward function, and
    training-step budget.}
    \label{tab:rl_hyperparams}
    \small
    \setlength{\tabcolsep}{6pt}
    \renewcommand{\arraystretch}{1.15}
    \begin{tabular*}{\textwidth}{@{\extracolsep{\fill}}l c c c@{}}
        \toprule
        \textbf{Hyperparameter} & \textbf{PPO} & \textbf{RPPO} & \textbf{TRPO} \\
        \midrule
        Policy architecture   & MLP & MLP + LSTM & MLP \\
        Requested timesteps   & $8 \times 10^5$ & $8 \times 10^5$ & $8 \times 10^5$ \\
        Actual timesteps      & 819,200 & 819,200 & 819,200 \\
        Parallel environments & 100 & 100 & 100 \\
        \(n_{\text{steps}}\)   & 256 & 256 & 256 \\
        Batch size            & 512 & 512 & 512 \\
        \(n_{\text{epochs}}\)  & 5 & 5 & --- \\
        Learning rate         & $3 \times 10^{-4}$ & $3 \times 10^{-4}$ & $3 \times 10^{-4}$ \\
        Discount \(\gamma\)   & 0 & 0 & 0 \\
        GAE \(\lambda\)       & 0.98 & 0.98 & 0.98 \\
        Clip range            & 0.2 & 0.2 & --- \\
        Entropy coefficient   & \(0.15\) & \(0.15\) & --- \\
        Target KL             & --- & --- & 0.01 \\
        Line search max steps & --- & --- & 15 \\
        Device                & CPU & CPU & CPU \\
        \bottomrule
    \end{tabular*}
\end{table}

\subsection{Baseline optimizers and screening protocols}
\label{sec:baselines_metrics}

\paragraph{Optimization baselines.}
We evaluated six baselines: random search; L-SHADE, implemented
through \texttt{mealpy}~\cite{van2023mealpy};
PSO~\cite{kennedy1995pso}; RUN~\cite{ahmadianfar2021run};
CMA-ES~\cite{hansen2001cmaes}; and Ax Bayesian optimization
using the BoTorch ecosystem~\cite{balandat2020botorch}.
All methods used the same encoded recipe space, action bounds,
and frozen surrogate.

For L-SHADE, PSO, and CMA-ES, a separate
Optuna~\cite{akiba2019optuna} study evaluated eight
hyperparameter trials per method. Each trial used two random
seeds and 500 candidate evaluations per seed. The tuning
objective was the mean of the best surrogate scores obtained
under the two seeds. Selected hyperparameters were fixed before
the main comparison. Baseline configurations are provided in
Appendix~\ref{app:baselines}.

\paragraph{Post-training evaluation budget.}
Each evaluation run used 3,000 candidate-scoring queries to the
frozen PCE surrogate. Baseline runs used 30 batches of 100
evaluations and were separately initialized with seeds 42,
12387, and 24732.

Each RL algorithm was evaluated under three inference seeds
using one frozen trained policy. Each inference run generated
3,000 candidates over 30 steps in 100 parallel environments.
Initial and auto-reset observations were excluded from the
candidate pools. For RPPO, the recurrent state was reset at
the start of each inference episode, including after each
auto-reset.

The evaluation budget counted candidate-scoring queries made
during these runs. RL training and baseline hyperparameter
tuning were excluded from this count.

\paragraph{Surrogate-objective summaries.}
For each run, we recorded the highest surrogate score observed
within the 3,000-query budget.
Table~\ref{tab:convergence} reports the mean and standard deviation
of the three run-level maxima for each method. The RL summaries
used runs with different inference seeds for a fixed trained
policy, whereas the baseline summaries used separately initialized
optimization runs.

For the optimization baselines, we also recorded the first
evaluation at which the surrogate score reached 95\% of that
run's final within-budget maximum. The target was calculated
separately for each run.

\paragraph{Shortlist construction and duplicate handling.}
For each RL policy and optimization baseline, we ranked the
candidates from the designated seed-42 run by surrogate-predicted
PCE and retained the 100 highest-scoring unique recipes for
neighborhood assessment.

RL shortlist construction included a raw-action duplicate check.
For all methods, recipe identity was checked after stoichiometric
normalization, so raw ratio vectors that encoded the same recipe
shared a single identity. The normalization-based check removed
no additional entries from the seed-42 RL candidate pools.

\paragraph{Candidate sets for downstream assessment.}
Neighborhood assessment used the 100-candidate shortlist from
each of the three RL policies and six optimization baselines.
These shortlists were assessed against the 890-record experimental
reference pool (Section~\ref{sec:neighborhood_methods}).

Composition-family matching used the full seed-42 RL candidate
pools: 3,000 PPO, 3,000 RPPO, and 3,000 TRPO recipes, for a total
of 9,000 recipes (Section~\ref{sec:openalex-method}).

\paragraph{Unselected sampling reference.}
We independently sampled 1,000 raw recipe vectors uniformly over
the encoded action variables. This Pure Random sample was assessed
without surrogate scoring or score-based selection. It served as
the unselected reference for neighborhood support and external
composition-family matching, supplied the denominator of the
neighborhood-support enrichment factor, and was used to construct
the sampling intervals in the rank-resolved figures. Pure Random
fractions were computed over all 1,000 sampled vectors; complete
shortlist fractions were computed over 100 candidates per method.

\subsection{Experimental-neighborhood scoring and enrichment}
\label{sec:neighborhood_methods}

\paragraph{Similarity to measured reference devices.}
We compared generated recipes with the 890-record experimental
reference pool using a weighted similarity metric. The metric
comprised 11 components: composition, four layer stacks, four
deposition procedures, band gap, and the quenching flag.

Composition similarity used an element-wise symmetric mean
absolute percentage error (SMAPE)-style score over the seven
represented species. Textual components used term
frequency--inverse document frequency (TF-IDF) cosine similarity,
with the TF-IDF models fitted on the reference pool. Band-gap
similarity used range-normalized absolute differences, and
quenching similarity used exact agreement.

For candidate $q$ and reference recipe $r$, aggregate similarity
was defined as
\begin{equation}
S(q,r)=
\frac{\sum_{c=1}^{11}W_c\,m_c(q,r)\,S_c(q,r)}
     {\sum_{c=1}^{11}W_c\,m_c(q,r)},
\label{eq:recipe_similarity}
\end{equation}
where $S_c(q,r)$ is the component similarity, $W_c$ is the
component weight, and $m_c(q,r)\in\{0,1\}$ indicates whether
the weight is included in the comparison. Similarity was set to
zero when the denominator was zero.

Component weights were derived from SHAP feature
importances~\cite{lundberg2017shap} in an auxiliary Random Forest
regressor~\cite{breiman2001randomforests}. The component formulas
and weights, including the species-level weights for composition,
are provided in Appendix~\ref{app:similarity}.

\paragraph{Missing descriptors.}
Missing textual entries contributed zero component similarity
while retaining their weights in the denominator. A textual
component was omitted if its entire column was absent from
either input data frame. For band gap and quenching, the
component weight was excluded when either value was missing at
comparison time. A reported false quenching flag was treated as
an observed value.

\paragraph{Neighborhoods and support categories.}
For each candidate $q$, we retained up to $k\in\{3,5,10\}$
reference neighbors with the highest similarities among those
satisfying $S(q,r)\geq\tau$, where $\tau=0.50$. This threshold
corresponded to the 75th percentile of pairwise similarities
in the reference pool. If fewer than $k$ eligible neighbors
existed, all eligible neighbors were retained.

Let $\mathcal{N}_k(q)$ denote the selected neighborhood. A
candidate was covered when $|\mathcal{N}_k(q)|\geq 1$. For a
covered candidate, the neighborhood score was the arithmetic
mean measured PCE:
\begin{equation}
\overline{y}_k(q)=
\frac{1}{|\mathcal{N}_k(q)|}
\sum_{r\in\mathcal{N}_k(q)}y_r,
\label{eq:neighborhood_mean}
\end{equation}
where $y_r$ is the measured PCE of reference device $r$.

Covered candidates were classified using quantiles of the
measured-PCE distribution in the reference pool:
\emph{Strong} if $\overline{y}_k(q)\geq Q_{0.85}$;
\emph{Positive} if $Q_{0.50}\leq\overline{y}_k(q)<Q_{0.85}$;
and \emph{Weak} if $\overline{y}_k(q)<Q_{0.50}$.
The thresholds were $Q_{0.50}=18.165\%$ and
$Q_{0.85}=22.86\%$. Candidates without an eligible neighbor
were classified as \emph{Uncovered}.

\paragraph{Neighborhood-supported hit rate.}
For a surrogate-ranked shortlist $\mathcal{L}_K$ containing
$K$ candidates, let $N_{\mathrm{covered}}$ be the number with
at least one eligible neighbor and
$N_{\mathrm{covered}\cap\mathrm{Strong},k}$ be the number
classified as Strong at maximum neighborhood size $k$.

Coverage and the conditional Strong fraction were defined as
\begin{equation}
C@K=\frac{N_{\mathrm{covered}}}{K},
\qquad
S_k@K=
\frac{N_{\mathrm{covered}\cap\mathrm{Strong},k}}
     {N_{\mathrm{covered}}},
\label{eq:coverage_conditional_strong}
\end{equation}
with $S_k@K$ undefined when $N_{\mathrm{covered}}=0$.
Coverage is the same for all tested values of $k$ because
coverage requires only one eligible neighbor and the similarity
threshold is fixed.

The neighborhood-supported hit rate was defined over all
shortlisted candidates:
\begin{equation}
\mathrm{HR}_{\mathrm{support},k}@K
=
\frac{N_{\mathrm{covered}\cap\mathrm{Strong},k}}{K}
=
(C@K)(S_k@K),
\label{eq:supported_hit_rate}
\end{equation}
where the product identity applies when coverage is nonzero.
The hit rate is zero when no candidates are covered.
Table~\ref{tab:coverage_adjusted_strong} reports the supported
hit rate at $K=100$.

For the same $K$, $k$, and reference criterion, relative
differences between methods were expressed as hit-rate ratios
when the denominator method had a nonzero hit rate. Absolute
differences were expressed in percentage points.

\paragraph{Enrichment relative to unselected sampling.}
We defined the enrichment factor as
\begin{equation}
\mathrm{EF}_k@K
=
\frac{\mathrm{HR}_{\mathrm{support},k}@K}{p_{0,k}},
\label{eq:enrichment_factor}
\end{equation}
where $p_{0,k}$ is the fraction of the 1,000 Pure Random
recipes that were both covered and Strong at maximum
neighborhood size $k$. At $k=5$, $p_{0,5}=29/1000=2.9\%$.
Both quantities in the ratio use all candidates in their
respective sets as the denominator. An enrichment factor of
one denotes a supported hit rate equal to that of the Pure
Random sample.

For each $K$, we used a fixed random seed to draw 1,000 subsets
of size $K$ from the Pure Random sample without replacement
within each draw. Enrichment, coverage, and mean neighborhood
PCE among covered recipes were computed on the same subset.
The enrichment calculations used $p_{0,k}$ from the full
Pure Random sample. For each metric, the reference interval
was defined by the 2.5th and 97.5th percentiles of the subset
statistics, with the median shown as the central line. Draws
with no covered recipes were omitted from the mean-PCE summary.
These intervals were conditional on the observed Pure Random
sample.

\paragraph{Complementary screening summaries.}
We reported the Strong, Positive, Weak, and Uncovered fractions
of each candidate set, together with the mean and standard
deviation of candidate-level neighborhood PCE among covered
candidates. Complete-shortlist summaries used $K=100$.
Rank-resolved analyses evaluated the top-$K$ prefixes of the
same shortlists at $k=5$ and expressed their supported hit
rates as enrichment factors
(Equation~\ref{eq:enrichment_factor}).

\paragraph{Internal assessment of the neighborhood signal.}
We applied the same neighborhood procedure to each reference
recipe, excluding its self-match. Spearman correlation quantified
the association between neighborhood-mean PCE and measured PCE
within the reference pool. Bootstrap confidence intervals,
PCE-label permutation controls, and random-neighbor controls
are described in Appendix~\ref{app:signal_validation}.

\subsection{External composition-family analysis}
\label{sec:openalex-method}

\paragraph{Literature retrieval.}
We constructed a separate composition reference pool using
OpenAlex~\cite{priem2022openalex}. The eleven search terms are
listed in Appendix~\ref{app:openalex-queries}. We restricted
queries to works published from 2024 onward and marked
\texttt{is\_oa=false} to distinguish the retrieval sources
from the open-access papers used for LLM extraction.
Duplicate works were removed by OpenAlex ID.

\paragraph{Composition extraction.}
Available abstracts were reconstructed from the
\texttt{abstract\_inverted\_index} field. Composition strings
were extracted from titles and abstracts using a pattern-based
parser. The parser expanded factored formulas, for example,
\[
\mathrm{(FAPbI_3)_{0.85}(MAPbBr_3)_{0.15}}
\rightarrow
\mathrm{FA_{0.85}MA_{0.15}PbI_{2.55}Br_{0.45}}.
\]
Only extracted compositions within the seven-species
representation were eligible, and works without an extractable
eligible composition were excluded.

\paragraph{Matching criterion.}
Matching used absorber composition for the 9,000 RL-generated
recipes and the 1,000 Pure Random recipes described in
Section~\ref{sec:baselines_metrics}. Each candidate composition
was compared with the retrieved reference compositions using
the composition-similarity component defined in
Appendix~\ref{app:similarity}.

The five nearest reference compositions were retained for
inspection. Classification was based on the highest similarity:
for candidate $q$, let
\[
b(q)=\max_{r\in\mathcal{R}_{\mathrm{comp}}}
S^{\mathrm{comp}}(q,r),
\]
where $\mathcal{R}_{\mathrm{comp}}$ is the retrieved composition
reference pool. A candidate was classified as matched when
$b(q)\geq\tau_{\mathrm{comp}}$. The matching rate was the
fraction of analyzed recipes that met this criterion.

We evaluated thresholds of $0.50$, $0.60$, $0.70$, $0.80$,
$0.85$, $0.90$, and $0.95$. After inspecting the results, we
selected $\tau_{\mathrm{comp}}=0.80$ because it gave the largest
difference in matching rates between the pooled RL-generated
recipes and the Pure Random sample among the evaluated
thresholds. The full threshold sensitivity analysis is provided
in Appendix~\ref{app:openalex-threshold}.

\section*{Data availability}

The processed datasets, reference-pool identifiers,
generated candidate recipes, and source data supporting
the figures and tables will be made available upon request.
The original Perovskite Database is described in
Ref.~\cite{jacobsson2022perovskite}.

\section*{Code availability}

The code for preprocessing, reward-model evaluation,
RL training, baseline optimization, and reference-based
analyses will be made available upon request.

\section*{Acknowledgments}
This work is supported by the Hong Kong Quantum AI Lab, AIR@InnoHK of the Hong Kong Government. 

\bibliography{sn-bibliography}

\clearpage

\appendix

\section{Data Preparation and Feature Representation}
\label{app:data}

This appendix describes how a raw entry in the Perovskite Database
is transformed into the recipe vector used by the reward model
and the RL agent: 23 dimensions for the reward model and
21 dimensions for the agent. The pipeline has three stages:
filtering and cleaning the raw database, selecting and encoding
the features, and constructing the final vector.

\subsection{Data preparation}
\label{app:data:prep}

The source dataset, reward-model eligibility criteria,
and recipe-level deduplication rule are described in
Section~\ref{sec:dataset}. The following paragraphs provide additional
details on reference-pool construction and missing
band gap handling.

\paragraph{Construction of the neighborhood reference pool.}
\label{app:reference-pool}

The reference-pool eligibility criteria, screening
rules, and final recipe-level deduplication key are
described in Section~\ref{sec:dataset}. Before the screening, a duplicate
reduction retained the highest-PCE record among
records identical in all remaining columns other
than PCE.

The final pool contained 890 records, with reported
PCE ranging from \(0.10\%\) to \(26.83\%\).
Its measured-PCE distribution supplied the
classification thresholds used in
Section~\ref{sec:neighborhood_methods}: a median of \(18.165\%\) and an
85th percentile of approximately \(22.86\%\).
Missing-value handling during similarity calculation
is described in Appendix~\ref{app:similarity}.

\paragraph{Band gap imputation.}
Approximately \(20.9\%\) of entries in the cleaned full dataset lack reported perovskite band gaps. To retain these records, we
used a separate TabPFN regressor to predict band gap
from the seven-dimensional composition vector
(Cs, FA, MA, Pb, Sn, Br, I), as described in
Appendix~\ref{app:bandgap}.
On its reported held-out test set, the band gap model
achieved an MAE of \(0.015\)\,eV and
\(R^2 = 0.979\).
Band gap also contributes a weight of 0.092 to the
similarity metric, based on SHAP importance in the
auxiliary Random Forest
(Table~\ref{tab:similarity_weights}).

For the five-fold PCE cross-validation described in
Section~\ref{sec:reward_model}, records with missing band gaps were retained during complete-case filtering of the
other columns. After material cleaning and complete-case
filtering excluding band gap, 34,081 records remained.
Applying the PCE $\geq 18\%$ criterion retained
4,545 records; recipe-level deduplication reduced this
to 1,862 records; and the publication-year criterion
($\geq 2016$), applied after deduplication, yielded
1,846 eligible records. Of these, 722 (39.1\%) had
missing band gaps; this rate is higher than the
20.9\% in the full cleaned dataset because the two datasets contain different
record subsets. Imputation was performed separately
within each cross-validation fold: a fresh band-gap
regressor was fitted using only training-fold records
with observed band gaps and was then applied to missing
values in both the training and validation sets.
Imputed values were therefore generated independently
for each fold rather than precomputed and shared
across folds.

When preparing the frozen reward model used for recipe optimization,
missing band gaps in all 1{,}846 records, in both the training and
held-out partitions, were imputed once with the saved band-gap surrogate
described in Appendix~\ref{app:bandgap} before the 70{:}30 split. The same
surrogate supplied composition-derived band gaps for generated recipes
(Section~\ref{sec:reward_model}).

\subsection{Feature selection and sublayer distribution}
\label{app:data:features}

The selected base features and the distinction between
agent-controlled recipe entries and additional
reward-model inputs are described in Section~\ref{sec:recipe_formulation}.
Here, we provide the sublayer-count statistics and
rare-category handling details underlying that
representation.

\paragraph{Sublayer distribution.}
Layer stack sequences are reported as ordered lists of materials
(e.g., \texttt{c-TiO2 | m-TiO2 | SnO2}). We split each sequence into positional
sublayers, keeping a fixed maximum number per stack.
Table~\ref{tab:sublayer_dist} reports the fraction of entries with \(1\),
\(2\), \(3\), or \(4\) or more sublayers for each of the four stack types.
Substrate and backcontact stacks almost always contain one or two sublayers (\(99.8\%\) combined), and HTL stacks contain a single sublayer in \(87.2\%\) of entries. Only the ETL stack has a non-negligible fraction of three-sublayer
entries (\(9.8\%\)) and a small tail at four or more (\(0.3\%\)). We
therefore keep up to two sublayers for the substrate, HTL, and backcontact
stacks, and up to three for the ETL stack. This limits the vector
dimensionality while retaining the sequences that appear in practice.

\begin{table}[htbp]
    \centering
    \caption{Sublayer-count distribution per stack type in the cleaned
    dataset. Percentages are computed over entries with a non-missing stack
    sequence for that field. The sublayer counts are used to choose the
    positional truncation length for each stack in the recipe vector.}
    \label{tab:sublayer_dist}
    \begin{tabular*}{\textwidth}{@{\extracolsep{\fill}}l cccc@{}}
        \toprule
        \textbf{Stack} & \textbf{1} & \textbf{2} & \textbf{3} & \textbf{\(\ge 4\)} \\
        \midrule
        Substrate   & \(4.5\%\)  & \(95.3\%\) & \(0.1\%\) & \(0.2\%\) \\
        ETL         & \(37.2\%\) & \(52.7\%\) & \(9.8\%\) & \(0.3\%\) \\
        HTL         & \(87.2\%\) & \(12.4\%\) & \(0.5\%\) & \(0.0\%\) \\
        Backcontact & \(95.3\%\) & \(4.5\%\)  & \(0.2\%\) & \(0.0\%\) \\
        \bottomrule
    \end{tabular*}
\end{table}

\paragraph{Rare category collapse.}
After positional splitting, many sublayer materials appear in only one
entry. We grouped every material that appears only once into a shared
\texttt{Other} category. This prevents the target encoder from assigning unique codes to singletons, thereby reducing the risk of overfitting. When a generated recipe decodes to \texttt{Other}, the specific
material cannot be recovered from the reward model's representation.
We therefore report such entries as \texttt{Other}.

\subsection{Vector encoding and feature transforms}
\label{app:data:vector}

The 10 agent-controlled base features expand into a \(21\)-dimensional vector
\(s = (s_1, \dots, s_{21})\) that serves as both the observation space and
the action space of the RL environment. Table~\ref{tab:vector_layout} lists
the position of each feature group. The layout matches the
\texttt{MultiDiscrete} action space defined in the environment.

When the reward model evaluates a recipe, two additional non-agent features
are appended: the derived band gap (computed from the composition by the
band gap surrogate) and the publication year (fixed at 2026 for all
generated recipes). The reward model input is therefore \(23\)-dimensional.

\begin{table}[htbp]
    \centering
    \caption{Layout of the 21-dimensional recipe vector (RL action and
    observation space). Two additional
    non-agent features (band gap, fixed publication year) are appended at
    reward-model evaluation time.}
    \label{tab:vector_layout}
    \begin{tabular*}{\textwidth}{@{\extracolsep{\fill}}c l l@{}}
        \toprule
        \textbf{Position} & \textbf{Feature} & \textbf{Values} \\
        \midrule
        \(1\)    & Substrate layer 1          & Categorical \\
        \(2\)    & Substrate layer 2          & Categorical \\
        \(3\)    & ETL layer 1                & Categorical \\
        \(4\)    & ETL layer 2                & Categorical \\
        \(5\)    & ETL layer 3                & Categorical \\
        \(6\)    & HTL layer 1                & Categorical \\
        \(7\)    & HTL layer 2                & Categorical \\
        \(8\)    & Backcontact layer 1        & Categorical \\
        \(9\)    & Backcontact layer 2        & Categorical \\
        \(10\)    & Perovskite deposition      & Categorical \\
        \(11\)   & ETL deposition             & Categorical \\
        \(12\)   & HTL deposition             & Categorical \\
        \(13\)   & Backcontact deposition     & Categorical \\
        \(14\)   & Cs ratio                   & \(0\text{--}100\) \\
        \(15\)   & FA ratio                   & \(0\text{--}100\) \\
        \(16\)   & MA ratio                   & \(0\text{--}100\) \\
        \(17\)   & Pb ratio                   & \(0\text{--}100\) \\
        \(18\)   & Sn ratio                   & \(0\text{--}100\) \\
        \(19\)   & Br ratio                   & \(0\text{--}100\) \\
        \(20\)   & I ratio                    & \(0\text{--}100\) \\
        \(21\)   & Quenching-induced cryst.   & \(0 / 1\) \\
        \bottomrule
    \end{tabular*}
\end{table}

\paragraph{Categorical features.}
The 13 categorical positions (indices \(1\text{--}13\)) are the layer
materials and deposition procedures. Each is encoded as an integer index
into the list of surviving categories for that feature, as learned during
preprocessing. Rare materials are collapsed into \texttt{Other} as described
above.

\paragraph{Composition.}
The perovskite composition is represented by a seven-dimensional vector of ratios
\((\mathrm{Cs}, \mathrm{FA}, \mathrm{MA}, \mathrm{Pb}, \mathrm{Sn},
\mathrm{Br}, \mathrm{I})\). Each entry is discretized to an integer in
\([0, 100]\). During a step, the environment normalizes the raw ratios so
that the A-site fractions sum to \(1\), the B-site fractions sum to \(1\),
and the X-site fractions sum to \(3\), which enforces the represented ABX$_3$
stoichiometry. For example,
MAPbI\(_3\) is encoded as \((0, 0, 100, 100, 0, 0, 100)\) before
normalization and \((0, 0, 1, 1, 0, 0, 3)\) after normalization.

\paragraph{Reward-model feature pipeline.}
Categorical features pass through three successive transforms before
reaching the reward model: (i) ordinal encoding, (ii) target encoding, and
(iii) a quantile transform. Numerical features (the seven normalized
composition ratios, publication year, band gap, and the quenching flag) are
not transformed. The RL environment operates on the ordinal indices
(Table~\ref{tab:vector_layout}); steps (ii) and (iii) are applied inside the
reward model whenever a recipe is evaluated.

\paragraph{Ordinal encoding.}
Each categorical feature (13 retained positions; layer materials and
deposition procedures) is first mapped to an integer index with
\texttt{OrdinalEncoder}. Categories that occur only once in the dataset are
collapsed into \texttt{Other} beforehand. Categories not in the fitted
vocabulary map to \(-1\). The vocabulary and the rare-category mapping are
determined from the full dataset before the train/test split. This applies to the frozen reward model only since we want the agent to have all options in the database during recipe optimization. It is fitted on the training fold only for the cross-validation. They use no
target information. The resulting indices define the \texttt{MultiDiscrete}
action space and carry no ordering.

\paragraph{Target encoding.}
Ordinal indices are then target-encoded
(\texttt{category\_encoders.TargetEncoder}, smoothing \(f=10.0\), minimum
samples per leaf \(k=5\)). For a category \(c\) with \(n_c\) training
records and mean PCE \(\bar y_c\), the encoded value is
\[
  \tilde x_c = \lambda(n_c)\,\bar y_c + \bigl(1-\lambda(n_c)\bigr)\,\bar y,
  \qquad
  \lambda(n) = \frac{1}{1+\exp\!\left(-(n-k)/f\right)},
\]
where \(\bar y\) is the global mean PCE of the training partition.
Categories absent from the training partition are encoded as \(\bar y\).
The encoder is fitted on the training partition only.
This allows the reward model to learn category-specific effects without
increasing the dimensionality of the feature space.

\paragraph{Quantile transform.}
The target-encoded categorical columns are mapped to a standard normal
distribution with \texttt{QuantileTransformer}
(\texttt{output\_distribution='normal'}, 1000 quantiles, fitted on the
training partition only). The transform is monotone, so it preserves the
ranking of categories by encoded value and changes only their scale. 

\section{LLM Extraction Pipeline}
\label{app:extraction}

To supplement the Perovskite Database with recent recipes, we extracted
synthesis procedures from recent open-access papers published after the
Perovskite Database's main collection period. We used the
\texttt{deepseek-v4-flash} model accessed through an API with a structured
prompt that instructs the model to return one JSON object per distinct
device reported in the paper. A total of 667 extracted recipes were retained after
filtering and deduplication and contributed to the
1,846-record high-PCE dataset. These records were
included before five-fold cross-validation and
before construction of the separate 70:30 split
used to fit the frozen reward model.

\paragraph{Extraction prompt.}
The system prompt is designed to enforce three requirements:
(i) one JSON object per distinct recipe reported in the paper;
(ii) a fixed set of keys matching the Perovskite Database schema;
and (iii) high-level deposition procedure names only, with
operational details such as spin speed, annealing temperature,
and solvent choice omitted. The full prompt is reproduced below.

\begin{quote}
\small\ttfamily
You are an expert data extractor for the Perovskite Database Project. Your
task is to extract experimental data from the provided scientific text and
output it in a strict JSON format.

CRITICAL INSTRUCTION: A single paper often describes MULTIPLE distinct
solar cell recipes or device variations (e.g., a ``control'' device vs.\ a
``modified/champion'' device, varying perovskite compositions, or different
transport layers). You must identify and extract EVERY distinct recipe
found in the text.

CRITICAL RULES:
\begin{enumerate}
    \item OUTPUT FORMAT: You must output a JSON ARRAY of objects. Even if
    only one recipe is found, output it as an array containing a single
    object.
    \item NO EXTRA KEYS: Every object in the array must ONLY contain the
    exact keys requested below.
    \item MISSING DATA: If a value is not explicitly mentioned for a
    specific recipe, output \texttt{null}.
    \item PEROVSKITE DATABASE FORMATTING:
    \begin{itemize}
        \item Stack sequences MUST be separated by a pipe character `` |
        '' (e.g., ``Glass | FTO | TiO2-c | TiO2-mp'').
        \item Deposition procedures MUST use ONLY high-level method names
        (e.g., ``Spin-coating'', ``Evaporation'', ``Sputtering''). DO NOT
        include rpm, time, temperature, solvents, or anti-solvents.
        \item Multiple distinct deposition steps are separated by `` >>
        '' (e.g., ``Spin-coating >> Evaporation'').
        \item Band gaps must be a float representing eV.
        \item Photovoltaic metrics (Voc, Jsc, FF, PCE) must be numbers,
        not strings. Drop the units.
        \item Boolean fields must be strictly \texttt{true} or
        \texttt{false}.
    \end{itemize}
\end{enumerate}

Output a JSON array where EACH object represents a distinct recipe and
contains EXACTLY these keys:
\begin{verbatim}
[
  {
    "Perovskite_composition_long_form": "Full chemical composition (e.g., MAPbI3)",
    "Perovskite_deposition_procedure": "High-level method ONLY. (e.g., 'Spin-coating' or 'Spin-
    coating >> Spin-coating'). NO details.",
    "Cell_stack_sequence": "Full sequence (e.g., Glass | FTO | SnO2 | Perovskite | 
    spiro-OMeTAD | Au)",
    "Substrate_stack_sequence": "e.g., Glass | FTO",
    "ETL_stack_sequence": "e.g., TiO2-c | TiO2-mp",
    "ETL_deposition_procedure": "High-level method ONLY",
    "HTL_stack_sequence": "e.g., spiro-OMeTAD",
    "HTL_deposition_procedure": "High-level method ONLY",
    "Backcontact_stack_sequence": "e.g., Au or Ag",
    "Backcontact_deposition_procedure": "High-level method ONLY",
    "Ref_publication_date": "YYYY-MM",
    "Perovskite_band_gap": float,
    "Perovskite_deposition_quenching_induced_crystallization": boolean (true if anti-solvent used),
    "JV_default_Voc": float,
    "JV_default_Jsc": float,
    "JV_default_FF": float,
    "JV_default_PCE": float,
    "JV_light_intensity": "e.g., 100 mW/cm2 or 1 Sun"
  }
]
\end{verbatim}
\end{quote}

\paragraph{Output schema.}
Each extracted recipe is a JSON object containing 18 keys.
These include the Perovskite Database fields corresponding to
the features listed in Table~\ref{tab:features}, as well as
additional fields required for preprocessing
(e.g., \texttt{JV\_light\_intensity}). Before merging, the extracted values are passed through the same
validation and cleaning pipeline as the database entries,
including stoichiometry checks and deduplication, with the
exception of the light-intensity filter. Only about 40\% of the raw
extracted records (6{,}230 of 15{,}642) reported a light intensity,
since many papers do not state it, so filtering on it would have
removed most of the extracted data.

\paragraph{Validation.}
The prompt includes an explicit formatting specification, and a
programmatic validator checks each response against the specification
(allowed procedure names, pipe-separated stacks, and numeric metrics). If a response fails validation, the model is queried again with the specific errors appended to the conversation, for up to three attempts in total. Fields that still fail validation after three attempts are set to \texttt{null} rather than retained as malformed data.

\section{Band Gap Surrogate Model}
\label{app:bandgap}

The band gap surrogate predicts the perovskite band gap from the
7-dimensional composition vector alone. It is used to impute missing band gap values in the training
set (Appendix~\ref{app:data:prep}) and to compute the derived
band gap feature during RL inference
(Appendix~\ref{app:data:vector}).

\paragraph{Architecture.}
We use a \texttt{TabPFNRegressor}~\cite{hollmann2025tabpfn} with the default configuration. TabPFN is
a prior-fitted transformer that performs in-context learning at inference
time; it requires no gradient-based training and no hyperparameter tuning,
which makes it well suited to the small imputation dataset. The same model
family is used for the PCE reward model.

\paragraph{Training data.}
The band gap model used the seven-dimensional
composition vector (Cs, FA, MA, Pb, Sn, Br, I)
as its only input. Records were deduplicated by
composition string, retaining the first occurrence,
and invalid composition encodings were removed.
The remaining records with observed band gaps
were randomly split into 80\% training and
20\% test sets using random seed 42.
The model was fitted on the training split.

\paragraph{Performance.}
Predictions were obtained using the TabPFN
predictive mean. On the held-out test set,
the model achieved an MAE of 0.015\,eV,
an MSE of 0.001016\,eV$^2$, and
\(R^2=0.979\).
This standalone band gap evaluation was separate
from PCE cross-validation, for which the band gap
model was refitted within each training fold.

\section{Reward Model Training}
\label{app:reward_model}

The PCE reward model is a \texttt{TabPFNRegressor} trained on the cleaned
recipe dataset described in Appendix~\ref{app:data}. This section describes
the training protocol, the cross-validation procedure, and the evaluation
metrics. Results are reported in Section~\ref{sec:reward_model_results}.

\subsection{Model configuration}

We use TabPFN with the default configuration. No hyperparameter tuning was
performed. TabPFN is a prior-fitted transformer that performs in-context
learning at inference time: it uses the training set as context rather than
fitting parameters through gradient descent. Consequently, task-specific gradient-based training
hyperparameters, such as the learning rate, batch size,
and number of epochs, do not need to be specified.
Input encoding and scaling are described in
Appendix~\ref{app:data:vector}. The same model family is used for the
band gap surrogate in Appendix~\ref{app:bandgap}.

\subsection{Residual and calibration analysis}
\label{app:residual_analysis}

\begin{figure}[h]
\centering
\includegraphics[width=0.9\textwidth]{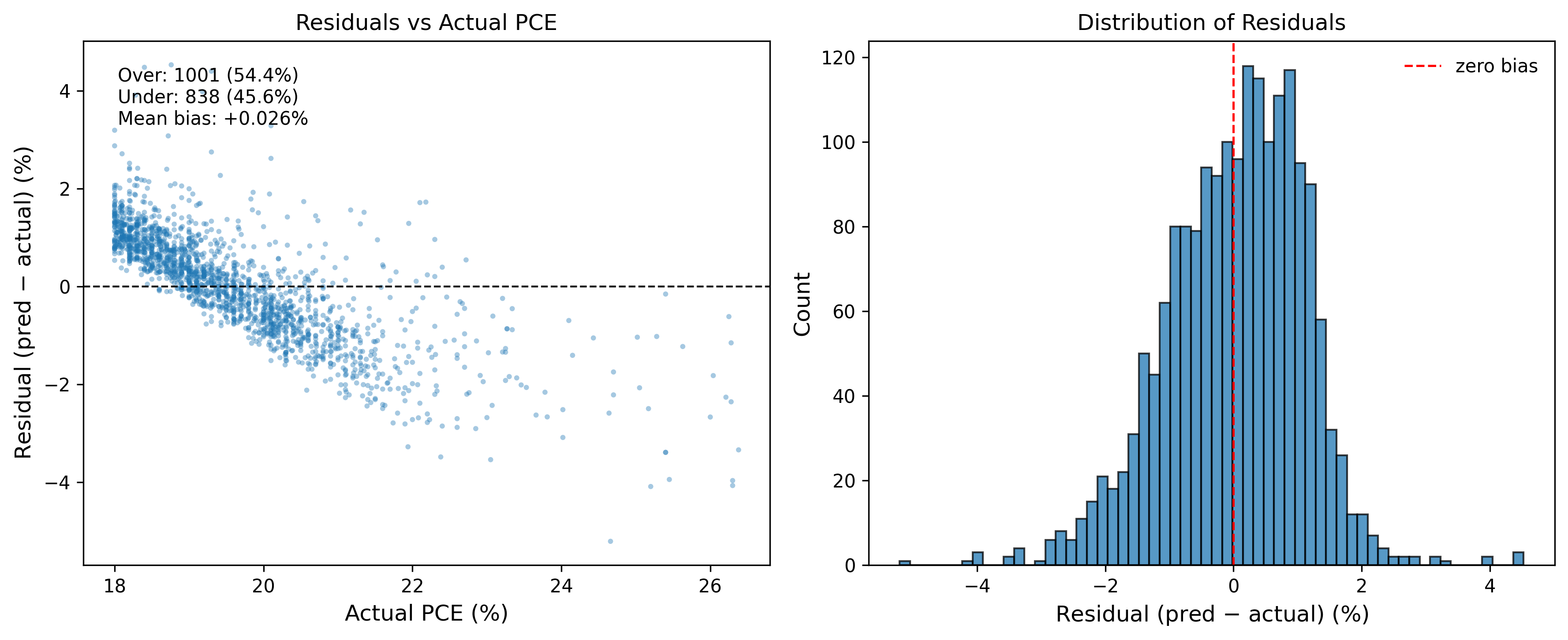}
\caption{Residual analysis across the five cross-validation folds.
\textbf{Left:} residuals \(r = \hat{y} - y\) as a function of actual PCE.
\textbf{Right:} distribution of residuals. The dashed line indicates zero
bias. The model overpredicts below \(20\%\) and underpredicts above \(20\%\),
with the underprediction growing at the high-PCE end.}
\label{fig:residual_analysis}
\end{figure}

Figure~\ref{fig:residual_analysis} shows the residual analysis, and Table~\ref{tab:bias_by_range} reports the prediction bias by actual PCE range. Across the 1,846 held-out predictions, \(53.1\%\) overestimate and
\(46.9\%\) underestimate the measured PCE, with a mean bias of
\(-0.006\%\) and a median bias of \(+0.127\%\). The residuals are approximately centered on zero, but their sign is not uniform across the PCE range. In the 18--20\% bin, the model overpredicts on average (mean bias \(+0.553\%\), 77.4\% are overestimates), whereas above 20\% it underpredicts, and the underprediction grows with PCE: \(-0.868\%\) in the 20--22\% range, \(-1.296\%\) in 22--24\%, and \(-2.551\%\) above 24\% (Table~\ref{tab:bias_by_range}). This trend is consistent with the reduced sample size at the high-efficiency end of the Perovskite Database: the model regresses toward the mean of the training distribution rather than extrapolating to the top of the range. 

The observed underprediction characterizes held-out experimental devices grouped by their measured PCE; it does not establish that recipes assigned high surrogate scores will outperform those scores experimentally. In particular, the analysis does not rule out reward-model exploitation when optimization proposes recipes outside the training distribution. The highest measured-PCE bin contains only 35 held-out samples, further limiting conclusions about predictive behavior in this region. The residual results therefore suggest a tendency to underestimate known high-performing devices, rather than validating the agent's highest-scoring proposals. Independent evaluation and experimental measurements are needed to determine whether these proposals represent genuine performance improvements.

\begin{table}[htbp]
\centering
    \caption{Prediction bias by actual PCE range, aggregated across the
        five cross-validation folds. Mean bias is the average of
        $\hat{y} - y$ over held-out records in each bin, in percentage
        points. A positive mean bias indicates overestimation.
        ``\% over'' and ``\% under'' are the fractions of records in the bin
        for which $\hat{y} > y$ and $\hat{y} < y$. Bins are pooled across
        folds; the pooled mean bias across all 1{,}846 records
        is the same as the count-weighted mean of the per-bin values.
        The \(<18\%\) bin is absent because the cleaned dataset is
        restricted to PCE \(\geq 18\%\).}
    \label{tab:bias_by_range}
    \begin{tabular*}{\textwidth}{@{\extracolsep{\fill}}l c c c c@{}}
    \toprule
    Actual PCE range (\%) & Count & Mean bias (\%) & \% over & \% under \\
    \midrule
    \(18\text{--}20\) & 1{,}184 & \(+0.553\) & 77.4 & 22.5 \\
    \(20\text{--}22\) & 552     & \(-0.868\) & 9.0  & 90.9 \\
    \(22\text{--}24\) & 75      & \(-1.296\) & 20 & 80.0 \\
    \(>24\)          & 35      & \(-2.551\) & 0  & 100 \\
    \bottomrule
    \end{tabular*}
\end{table}

\subsection{Input features}

The reward model takes as input all \(23\) features: the \(21\) agent-controlled recipe features
(the recipe vector), the derived band gap, and a publication year, which is fixed during RL training.
No feature-selection step is applied inside the reward model. The SHAP~\cite{lundberg2017shap}
importances reported in Table~\ref{tab:similarity_weights} are computed
post-hoc using a random forest model~\cite{breiman2001randomforests} to weight the similarity metric
(Appendix~\ref{app:similarity}), not to prune the reward-model input.

\subsection{Cross-validation protocol}

We evaluated the reward model separately on the full
cleaned dataset and the high-PCE subset using five-fold
stratified cross-validation. The high-PCE
evaluation contained 1,846 records and used 20 quantile
bins of measured PCE for stratification, with shuffling
and random seed 50. Validation-fold sizes were 370,
369, 369, 369, and 369; the corresponding training-fold
sizes were 1,476, 1,477, 1,477, 1,477, and 1,477.
Every eligible record was evaluated exactly once. The high-PCE subset
contained 1,846 recipes in total, and its out-of-fold
predictions were used for the residual and
ranking analyses reported in Appendices~\ref{app:residual_analysis}
and \ref{app:surrogate-ranking}.

For each fold, preprocessing was performed as follows:

\begin{itemize}
    \item \textbf{Band-gap imputation.}
    A fresh TabPFN regressor was fitted using the seven
    composition features from training-fold records
    with observed band gaps. It was then used to impute
    missing band gaps in both the training and validation
    sets, leaving observed values unchanged. Across
    folds 1--5, the numbers of observed training band
    gaps were 894, 899, 901, 895, and 907, respectively.
    The numbers of imputed validation band gaps were
    140, 144, 146, 140, and 152, summing to the 722
    originally missing values.

    \item \textbf{Rare-category handling.}
    Categories appearing only once in a
    training-fold categorical feature were mapped to
    \texttt{Other}. The same mapping was applied to
    the corresponding validation feature.

    \item \textbf{Ordinal encoding.}
    Category mappings were learned from the training
    fold. Previously unseen validation categories
    received the code $-1$.

    \item \textbf{Target encoding.}
    The target encoder was fitted using training-fold
    features and PCE targets, with smoothing 10.0 and
    \texttt{min\_samples\_leaf=5}. The fitted mapping
    was then applied to the validation fold.

    \item \textbf{Categorical-feature transformation.}
    A quantile transformer with a normal output
    distribution was fitted to the target-encoded
    categorical features in the training fold and
    applied to the corresponding validation features.
\end{itemize}
A separate PCE model was fitted in each fold using
the same fixed set of 23 features. Predictions were
stored with their original record identifiers to verify
that every eligible record appeared in exactly one
validation fold. The source dataset remained unimputed
throughout cross-validation.

For subsequent recipe optimization, the PCE reward model and its fitted
preprocessing transforms used the 1{,}292-record training partition
described in Section~\ref{sec:reward_model}. Before the split, missing
band gaps in all records were imputed once using the separately fitted,
saved band-gap surrogate described in Appendix~\ref{app:bandgap}, rather
than refitted per fold as in the cross-validation. The PCE model,
preprocessing transforms, and saved band-gap surrogate were then held fixed.

\subsection{Metrics}
\label{app:metrics}
For each of the five folds, we record the following regression metrics:

\begin{itemize}
    \item \textbf{MAE}: mean absolute error in percentage points of PCE.
    \item \textbf{RMSE}: root mean squared error in percentage points of PCE.
    \item \textbf{\(R^2\)}: coefficient of determination on the validation fold.
\end{itemize}

Regression results are reported in Table~\ref{tab:reward} as mean \(\pm\) standard
deviation across the five folds, on both the full cleaned dataset and
the high-PCE subset (\(\mathrm{PCE} \geq 18\%\), publication year
\(\geq 2016\)). The mean fold-level results
were $R^2 = 0.4143 \pm 0.0767$,
MAE $= 0.8667 \pm 0.0212$ percentage points,
and RMSE $= 1.1183 \pm 0.0420$ percentage points.
Metrics calculated from all pooled out-of-fold
predictions were $R^2 = 0.4250$,
MAE $= 0.8667$ percentage points,
and RMSE $= 1.1191$ percentage points.
Pooled metrics and mean fold-level metrics are
distinct summaries and need not be identical.

To assess candidate-prioritization performance, we additionally calculate
the following metrics for our surrogate on the high-PCE subset, separately
within each validation fold:

\begin{itemize}
    \item \textbf{Spearman rank correlation}: rank correlation between
    predicted and measured PCE across the validation recipes.

    \item \textbf{Top-10 measured PCE}: mean measured PCE of the ten
    recipes with the highest predicted scores. If predictions tie at
    the selection cutoff, we calculate the expected mean under uniform
    selection among the tied recipes.

    \item \textbf{Random-selection reference}: mean measured PCE of the
    entire validation fold, which equals the expected mean PCE of ten
    recipes selected uniformly at random without replacement from that
    fold.

    \item \textbf{Top-10 selection gain}: difference between the top-10
    measured PCE and the random-selection reference, expressed in
    percentage points.
\end{itemize}

Ranking and candidate-selection results are reported in Appendix~\ref{app:surrogate-ranking}
as mean \(\pm\) standard deviation across the five folds. Selection is
performed within each fold rather than on pooled predictions from
different fitted models. These metrics assess prioritization of held-out
experimental recipes, not the performance of optimizer-generated recipes.

We do not report per-fold results in the main text; the per-fold metrics are reported in Appendix~\ref{app:surrogate-ranking}.

\section{Additional RL Implementation Details}
\label{app:hyperparams}

The RL implementations, training budgets, and
hyperparameters are described in Section~\ref{sec:rl_agent} and
Table~\ref{tab:rl_hyperparams}.
This appendix provides additional details on the episode termination, composition
normalization, and training diagnostics.

\paragraph{Episode structure.}
Episodes terminate after 200 steps or earlier if the predicted PCE changes
by less than \(0.0001\) over 15 consecutive steps. No per-step bonus or early-termination bonus is applied;
the only reward is the differential PCE improvement
\(r_t = R(s_{t+1}) - R(s_t)\).

\paragraph{Composition normalization and zero-sum fallback.}
The composition action is a discrete 7-vector of raw ratios. The environment
normalizes these ratios so that the resulting composition satisfies
\(\mathrm{Cs} + \mathrm{FA} + \mathrm{MA} = 1\),
\(\mathrm{Pb} + \mathrm{Sn} = 1\), and
\(\mathrm{Br} + \mathrm{I} = 3\), which enforces the \(\mathrm{ABX_3}\)
stoichiometry constraint. If all raw entries within a sublattice group are zero, that
group is assigned equal proportions and scaled to its required
stoichiometric total. The resulting fallback values are
$(1/3,\,1/3,\,1/3)$ for $(\mathrm{Cs},\,\mathrm{FA},\,\mathrm{MA})$,
$(1/2,\,1/2)$ for $(\mathrm{Pb},\,\mathrm{Sn})$, and
$(3/2,\,3/2)$ for $(\mathrm{Br},\,\mathrm{I})$.
This fallback ensures that normalization is well-defined
even for all-zero sublattice groups. The normalization
procedure is implemented as a projection inside the
environment's step function and does not require action masking.

\subsection{Training diagnostics}
\label{app:training_diagnostics}

PPO and RPPO maintain \texttt{approx\_kl} between
\(0.03\) and \(0.37\) and \texttt{clip\_fraction} between \(0.40\) and
\(0.64\). These values indicate that a substantial fraction of probability
ratios are clipped, so the updates are not uniformly small; PPO does not
impose a hard KL constraint in our configuration. TRPO's
\texttt{kl\_divergence\_loss} stays at \(0.0038\)--\(0.0092\), below its
target KL of \(0.01\), indicating that the line search keeps each update
within the trust region rather than approaching or exceeding the target. The order-of-magnitude difference between the two KL
quantities is expected because TRPO explicitly enforces a trust region while
PPO relies on ratio clipping. Entropy loss increases for PPO and RPPO
(\(-71.8 \rightarrow -55.1\) for PPO and \(-71.8 \rightarrow -51.8\) for
RPPO), which corresponds to
decreasing policy entropy and increasingly deterministic policies. TRPO does
not log entropy loss in our runs. Overall, PPO and RPPO converge to more deterministic policies, whereas TRPO takes smaller, trust-region-limited steps.

\section{Learning Curves and Discount-Factor Ablation}
\label{app:gamma}

\begin{figure}[H]
    \centering
    \includegraphics[width=\textwidth]{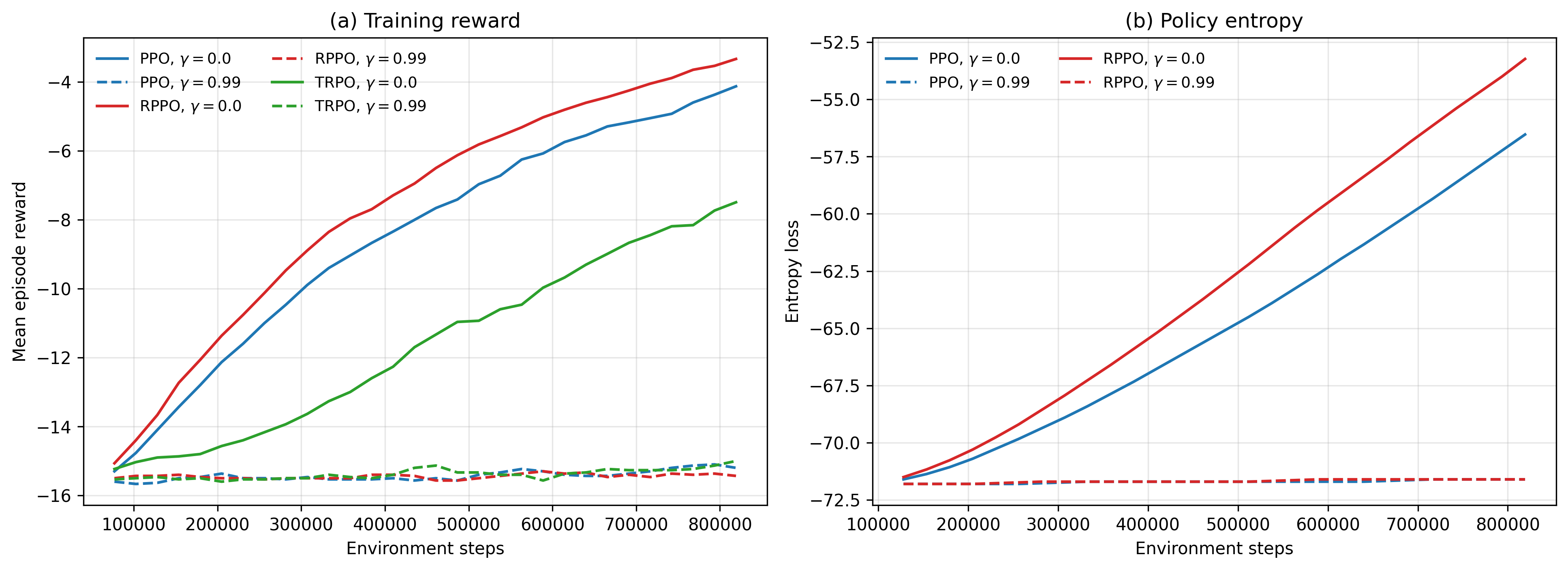}
    \caption{Learning dynamics under different discount factors.
    (a) Mean episode reward for PPO, RPPO, and TRPO with
    \(\gamma=0\) (solid) and \(\gamma=0.99\) (dashed).
    (b) Entropy loss for PPO and RPPO; TRPO does not log this
    quantity in the implementation used here.
    Within each algorithm, the two discount-factor settings use
    otherwise identical hyperparameters and a requested training
    budget of $8 \times 10^5$ environment steps, resulting in
    819,200 actual steps per run.
    Each trajectory represents one training run with
    \(100\) parallel environments.
    Each rollout comprises \(100 \times 256 = 25{,}600\)
    environment steps. Curves are smoothed using a
    three-point moving average.
    With \(\gamma=0.99\), rewards remain near the initial
    random-policy level, whereas \(\gamma=0\) produces
    substantial improvement.
    Increasing entropy loss in panel~(b) corresponds to
    decreasing policy entropy.}
\label{fig:gamma_ablation}
\end{figure}

\paragraph{Experimental setup.}
We compared \(\gamma = 0\) and \(\gamma = 0.99\) for PPO,
RPPO, and TRPO under otherwise identical hyperparameters and
an \(8 \times 10^{5}\)-step training budget.
Each run used \(100\) parallel environments and \(256\) steps
per environment per rollout.
Table~\ref{tab:gamma_ablation} summarizes the initial and final
mean episode rewards, entropy losses where available, and
value-function explained variances.
Figure~\ref{fig:gamma_ablation}(a) shows the corresponding
reward trajectories; the solid curves also provide the learning
curves for the \(\gamma = 0\) agents discussed in
Section~\ref{sec:rl_results}.
Figure~\ref{fig:gamma_ablation}(b) shows entropy loss for
PPO and RPPO. TRPO does not log entropy loss in the
\texttt{sb3-contrib} implementation used here.

\begin{table}[htbp]
    \centering
    \caption{Discount-factor ablation. Values are reported as initial $\to$ final
    for each training run, with a requested budget of $8\times10^{5}$ environment
    steps and an actual total of 819{,}200 steps. Mean episode reward is based on
    the differential predicted PCE, expressed in percentage points. Because the
    reset state is not evaluated by the surrogate, its reference value is set to
    the constant offset $R(s_0)=35$ (Section~\ref{sec:rl_agent}), so the return
    of an episode is $R(s_T)-35$ and the reported rewards are relative to this
    offset. For example, a final mean reward of $-3.1$ corresponds to an end-state
    surrogate score of about $31.9$. Entropy loss is the logged negative policy
    entropy, not a PCE quantity; increasing entropy loss corresponds to decreasing
    policy entropy. Explained variance is dimensionless. TRPO does not log entropy
    loss in the implementation used here, so the corresponding entries are marked
    ``---''. Each row summarizes one training run.}
    \label{tab:gamma_ablation}
    \renewcommand{\arraystretch}{1.15}
    \begin{tabular*}{\textwidth}{@{\extracolsep{\fill}}l l c c c@{}}
        \toprule
        \textbf{Algorithm} & \textbf{Discount}
            & \textbf{Mean episode reward}
            & \textbf{Entropy loss}
            & \textbf{Explained variance} \\
        \midrule
        \multirow{2}{*}{PPO}
            & \(\gamma = 0\)
            & \(-15.7 \to -3.9\)
            & \(-71.8 \to -55.1\)
            & \(0.02 \to 0.50\) \\
            & \(\gamma = 0.99\)
            & \(-15.2 \to -15.4\)
            & \(-71.8 \to -71.6\)
            & \(0.01 \to 0.92\) \\
        \midrule
        \multirow{2}{*}{RPPO}
            & \(\gamma = 0\)
            & \(-15.6 \to -3.1\)
            & \(-71.8 \to -51.8\)
            & \(0.00 \to 0.48\) \\
            & \(\gamma = 0.99\)
            & \(-15.5 \to -15.4\)
            & \(-71.8 \to -71.6\)
            & \(0.01 \to 0.94\) \\
        \midrule
        \multirow{2}{*}{TRPO}
            & \(\gamma = 0\)
            & \(-15.5 \to -7.1\)
            & ---
            & \(0.01 \to 0.36\) \\
            & \(\gamma = 0.99\)
            & \(-15.7 \to -14.9\)
            & ---
            & \(0.01 \to 0.92\) \\
        \bottomrule
    \end{tabular*}
\end{table}


\paragraph{Observed training behavior.}
With \(\gamma = 0.99\), mean episode reward remained near
the initial random-policy level for all three algorithms.
Entropy loss also changed little for PPO and RPPO,
from approximately \(-71.8\) to \(-71.6\).
Nevertheless, value-function explained variance reached
\(0.92\)--\(0.94\).
Thus, high explained variance did not coincide with
substantial policy improvement in these runs.

With \(\gamma = 0\), all three algorithms improved
substantially, reaching final mean episode rewards of
approximately \(-3.9\) for PPO, \(-3.1\) for RPPO, and
\(-7.1\) for TRPO.
Entropy loss increased to \(-55.1\) for PPO and \(-51.8\)
for RPPO, corresponding to decreasing policy entropy.
Although explained variance was lower than with
\(\gamma = 0.99\), policy performance improved.
These observations support the choice of \(\gamma = 0\)
for the main experiments, but do not establish the
cause of the difference on their own.

\paragraph{Approximate interpretation.}
The observed behavior is consistent with attenuation of
the action-dependent temporal-difference signal under
a high discount factor.
To illustrate this mechanism, consider a near-uniform,
state-independent policy, and let
\(\mu = \mathbb{E}_{a \sim \pi}[R(a)]\) denote its mean
predicted PCE.
Ignoring episode-boundary effects, the value function
under this approximation is
\(V^\pi(s) \approx \mu - R(s)\):
the expected immediate reward is \(\mu - R(s)\),
whereas subsequent differential rewards have zero
expectation under the fixed state-independent policy.

For a non-terminal transition, substituting this
approximation into the temporal-difference residual gives
\begin{equation}
\begin{aligned}
\delta_t
&= r_t + \gamma V^\pi(s_{t+1}) - V^\pi(s_t) \\
&\approx R(a_t) - R(s_t)
  + \gamma\bigl[\mu - R(a_t)\bigr]
  - \bigl[\mu - R(s_t)\bigr] \\
&= (1-\gamma)\bigl[R(a_t)-\mu\bigr],
\end{aligned}
\end{equation}
where \(s_{t+1}=a_t\).

Under this approximation, the action-dependent component
of the raw temporal-difference signal is scaled by
\(1-\gamma\). At \(\gamma=0.99\), its magnitude is therefore
one hundredth of that at \(\gamma=0\), for the same
\(R(a_t)-\mu\).
However, this does not imply that the actual policy
gradient or policy update is one hundredth as large.
Advantage normalization, value-estimation error,
multi-step advantage estimation, episode termination,
and other implementation details also affect learning.
The derivation therefore provides a possible mechanism,
rather than a complete causal explanation of the
observed training behavior.

High explained variance measures the fit of the value
function to its training targets; it does not directly
measure the quality of the advantages used for policy
improvement. In particular, the approximate value
function \(\mu-R(s)\) remains state-dependent even when
the action-dependent temporal-difference signal is small.
Setting \(\gamma=0\) removes the bootstrapped future-value
term and makes the value target the immediate
differential reward. In the reported runs, this setting
yielded substantial policy improvement across all three
algorithms.

\section{Baseline Hyperparameters}
\label{app:baselines}

Baselines are implemented with mealpy, cma, and Ax;
Optuna is used only for hyperparameter tuning.
L-SHADE, PSO, and CMA-ES were each tuned using a
short Optuna study with a TPE sampler, eight trials per method,
two random seeds per trial, and 500 candidate evaluations per
seed. The tuning objective was the mean best surrogate score
across the two seeds. Selected hyperparameters were frozen
for the main comparison and are listed in
Table~\ref{tab:baseline_hyperparams}. Random
search, RUN, and Ax BO did not undergo this tuning stage;
their configurations are specified in Table~\ref{tab:baseline_hyperparams}.

\begin{table}[htbp]
    \centering
    \caption{Hyperparameters for the six optimization baselines. Values
    marked ``default'' use the library's recommended settings.}
    \label{tab:baseline_hyperparams}
    \begin{tabular*}{\textwidth}{@{\extracolsep{\fill}}l l l@{}}
        \toprule
        \textbf{Method} & \textbf{Tuned parameters} & \textbf{Values} \\
        \midrule
        Random Search      & --- & --- \\
        L-SHADE            & \(\mu_F\), \(\mu_{CR}\) & \(0.766\), \(0.270\) \\
        PSO                & \(c_1\), \(c_2\), \(w\) & \(2.197\), \(1.312\), \(0.225\) \\
        RUN                & --- & default \\
        CMA-ES             & \(\sigma_0\) & \(1.237\) \\
        Ax BO              & --- & default \\
        \bottomrule
    \end{tabular*}
\end{table}

All baselines use the same discrete recipe encoding and
action-dimension bounds as the RL agents. Each baseline is
evaluated using 30 batches of 100 evaluations per run,
with three independently seeded runs
(seeds \(42\), \(12387\), and \(24732\)). The inference seeds for RL agents are the same.

\section{Held-out ranking evaluation of the surrogate}
\label{app:surrogate-ranking}

To assess the surrogate's suitability for candidate prioritization,
we evaluated ranking performance within each held-out
cross-validation fold. We calculated Spearman's rank correlation
between predicted and measured PCE and selected the ten recipes
with the highest predicted scores in each validation fold.
We then calculated their mean measured PCE and compared it with
the expected mean PCE under uniform random selection from the
same validation pool. This random-selection expectation equals
the mean measured PCE of the validation pool. Selection gain
was defined as the difference between the top-ten mean measured
PCE and this random-selection expectation, expressed in
percentage points.

The model achieved a Spearman rank correlation of
\(0.436 \pm 0.037\). The mean measured PCE of the top-ten
selections was \((23.737 \pm 0.853)\%\), compared with
\((19.684 \pm 0.032)\%\) under uniform random selection. The
corresponding selection gain was \(4.053 \pm 0.822\) percentage
points and was positive in all five folds.
Aggregate values are reported as mean \(\pm\) standard deviation across
folds. Per-fold results are provided in
Table~\ref{tab:surrogate-ranking}.

These results support the surrogate's ability to prioritize
higher-performing recipes within the evaluated dataset.
They do not establish its ranking accuracy for optimizer-generated
recipes outside that distribution or its superiority over
alternative surrogate models.

\begin{table}[htbp]
    \centering
    \caption{Held-out regression and ranking performance across the five
    cross-validation folds. MAE, RMSE, and $R^2$ are computed on the
    validation fold. Spearman $\rho$ is the rank correlation between
    predicted and measured PCE. Top-ten PCE is the mean measured PCE of
    the ten recipes with the highest predicted scores in each validation
    fold; Random PCE is the expected mean under uniform random selection
    from that fold; Gain is the difference in percentage points (pp).
    The final row reports the mean and, on the second line, the
    standard deviation across folds.}
    \label{tab:surrogate-ranking}
    \footnotesize
    \setlength{\tabcolsep}{4pt}
    \renewcommand{\arraystretch}{1.05}
    \begin{tabular*}{\textwidth}{@{\extracolsep{\fill}}l r cccc ccc@{}}
        \toprule
        Fold & $n$ & MAE & RMSE & $R^2$
             & \shortstack{Spearman\\$\rho$}
             & \shortstack{Top-ten\\PCE (\%)}
             & \shortstack{Random\\PCE (\%)}
             & \shortstack{Gain\\(pp)} \\
        \midrule
        1 & 370 & 0.864 & 1.146 & 0.504 & 0.457 & 24.680 & 19.721 & 4.959 \\
        2 & 369 & 0.895 & 1.160 & 0.343 & 0.410 & 23.588 & 19.671 & 3.917 \\
        3 & 369 & 0.865 & 1.095 & 0.306 & 0.382 & 22.264 & 19.639 & 2.626 \\
        4 & 369 & 0.831 & 1.047 & 0.444 & 0.487 & 23.674 & 19.669 & 4.005 \\
        5 & 369 & 0.879 & 1.144 & 0.475 & 0.441 & 24.479 & 19.721 & 4.758 \\
        \midrule
        \shortstack{Mean\\$\pm$SD}
             &
             & \shortstack{$0.867$\\$\pm0.021$}
             & \shortstack{$1.118$\\$\pm0.042$}
             & \shortstack{$0.414$\\$\pm0.077$}
             & \shortstack{$0.436$\\$\pm0.037$}
             & \shortstack{$23.737$\\$\pm0.853$}
             & \shortstack{$19.684$\\$\pm0.032$}
             & \shortstack{$4.053$\\$\pm0.822$} \\
        \bottomrule
    \end{tabular*}
\end{table}
\section{Similarity Metric}
\label{app:similarity}

This appendix describes the similarity function used throughout the paper
to compare generated recipes against real recipes in the Perovskite
Database. The metric reflects the heterogeneous nature of a recipe,
which comprises a composition, layer stacks, deposition
procedures, a band gap, and a processing flag.
A recipe that matches a real device in composition but uses an incompatible
electron transport layer, for example, is not a strong match. The metric therefore decomposes the recipe into feature
groups, computes a per-group similarity, and aggregates
the groups using weights derived from
SHAP~\cite{lundberg2017shap} feature importances computed
on a separately trained random forest
regressor~\cite{breiman2001randomforests}.

\subsection{Component decomposition}

A recipe is partitioned into \(C = 11\) components, listed in
Table~\ref{tab:similarity_components}. Each component groups one or more
features of the reward model, and each is assigned a similarity function
appropriate to its type. A recipe is thus represented as
\(x = (x_1, \dots, x_C)\), where \(x_c\) is the value of component \(c\).

\begin{table}[htbp]
    \centering
    \caption{Components of the similarity metric. Each component groups one
    or more features of the reward model and is scored by a similarity
    function appropriate to its data type.}
    \label{tab:similarity_components}
    \footnotesize
    \setlength{\tabcolsep}{4pt}
    \renewcommand{\arraystretch}{1.1}
    \begin{tabularx}{\textwidth}{@{} l
        >{\raggedright\arraybackslash}X
        >{\raggedright\arraybackslash}p{2.4cm} @{}}
        \toprule
        \textbf{Component} & \textbf{Feature(s)} & \textbf{Similarity} \\
        \midrule
        Composition
            & \texttt{comp\_Cs}, \texttt{comp\_FA}, \texttt{comp\_MA},
              \texttt{comp\_Pb}, \texttt{comp\_Sn},
              \texttt{comp\_Br}, \texttt{comp\_I}
            & Per-element SMAPE \\
        Substrate stack
            & \texttt{sub\_layer\_1}, \texttt{sub\_layer\_2}
            & TF-IDF cosine \\
        ETL stack
            & \texttt{etl\_layer\_1}, \texttt{etl\_layer\_2},
              \texttt{etl\_layer\_3}
            & TF-IDF cosine \\
        HTL stack
            & \texttt{htl\_layer\_1}, \texttt{htl\_layer\_2}
            & TF-IDF cosine \\
        Backcontact stack
            & \texttt{back\_layer\_1}, \texttt{back\_layer\_2}
            & TF-IDF cosine \\
        Perovskite deposition
            & \texttt{Perovskite\_deposition\_procedure}
            & TF-IDF cosine \\
        ETL deposition
            & \texttt{ETL\_deposition\_procedure}
            & TF-IDF cosine \\
        HTL deposition
            & \texttt{HTL\_deposition\_procedure}
            & TF-IDF cosine \\
        Backcontact deposition
            & \texttt{Backcontact\_deposition\_procedure}
            & TF-IDF cosine \\
        Band gap
            & \texttt{Perovskite\_band\_gap}
            & Range-normalized \\
        Quenching
            & \texttt{Perovskite\_deposition\_\allowbreak
                    quenching\_\allowbreak
                    induced\_\allowbreak
                    crystallization}
            & Exact match \\
        \bottomrule
    \end{tabularx}
\end{table}

\subsection{Per-component similarity}
\paragraph{Composition.}
Each composition string---whether taken from a database record, a generated
recipe, or a retrieved OpenAlex reference---is parsed into the seven-species
vector
\begin{equation}
\mathcal{E}
= \{\mathrm{Cs},\, \mathrm{FA},\, \mathrm{MA},\, \mathrm{Pb},\,
   \mathrm{Sn},\, \mathrm{Br},\, \mathrm{I}\},
\label{eq:modelled_species}
\end{equation}
and let $\hat{x}_e$ denote the normalized fraction of species $e$ (so that
$\sum_{e \in \mathcal{E}} \hat{x}_e = 1$). Two composition vectors are
compared element-wise using a symmetric mean absolute percentage error
(SMAPE)-style similarity:
\[
    s^{\text{comp}}_e(q, r)
    = 1 - \frac{\lvert \hat{x}_e(q) - \hat{x}_e(r) \rvert}
             {\lvert \hat{x}_e(q) \rvert + \lvert \hat{x}_e(r) \rvert + \varepsilon},
\]
with \(\varepsilon = 10^{-9}\) to avoid division by zero. The score lies in
\([0, 1]\): \(1\) when both fractions are equal, \(0\) when one is zero and
the other is not. Because the denominator scales with the magnitudes involved, the metric
penalizes \emph{relative} movement:
$\mathrm{Cs}_{0.1} \to \mathrm{Cs}_{0.2}$ (a doubling) is scored as a
substantially larger change than $\mathrm{Cs}_{0.3} \to \mathrm{Cs}_{0.4}$,
even though both add the same absolute amount.

Compositions containing a species outside $\mathcal{E}$---such as Rb, K, Na,
Ge, or Cl---cannot be represented in this vocabulary and are excluded from
the reference pool before matching. The component-level composition similarity is the
SHAP-weighted average over the seven modelled species:
\[
    S^{\text{comp}}(q, r)
    = \frac{\sum_{e \in \mathcal{E}} w_e \, s^{\text{comp}}_e(q, r)}
           {\sum_{e \in \mathcal{E}} w_e},
\]
where \(w_e\) is the SHAP importance of the feature
corresponding to constituent \(e\), computed using a
separate Random Forest model trained on the same features
(e.g., \(w_{\text{FA}} =
\text{SHAP}(\texttt{comp\_FA})\)). Constituents with high importance,
such as FA and MA in the auxiliary Random Forest, contribute
more strongly to the score. This weighting emphasizes recipe features with
greater predictive importance in that model.

\paragraph{Text columns.}
Each of the eight textual components---four stacks and
four deposition procedures---is vectorized using a
TF-IDF model fitted on the reference pool. The resulting
vectors are compared using cosine similarity:
\[
    S^{\text{text}}_c(q, r)
    = \cos\!\left(\text{TF-IDF}_c(q), \, \text{TF-IDF}_c(r)\right).
\]
Identical nonzero TF--IDF vectors have similarity 1, while
vectors with no shared retained features have similarity 0.
Missing strings are represented by zero vectors and contribute
zero similarity, including when both strings are missing.

\paragraph{Band gap.}
Band gap is compared by range-normalized absolute difference:
\[
    S^{\text{bg}}(q, r)
    = \max\!\left(0, 1 - \frac{\lvert g(q) - g(r) \rvert}
                          {\max_{r'} g(r') - \min_{r'} g(r')}\right),
\]
where the range is computed over the reference pool. This maps a deviation
of one full range width to a similarity of \(0\), and identical band gaps
to \(1\).

\paragraph{Boolean.}
The quenching-induced crystallization flag is a binary feature. Its
similarity is \(1\) if \(q\) and \(r\) agree and \(0\) otherwise.

\subsection{Component weighting by SHAP importance}

Each component \(c\) is assigned a weight \(W_c\) equal to the sum of SHAP
importances of its constituent features:
\[
    W_c = \sum_{f \in \mathcal{F}_c} \text{SHAP}(f),
\]
where \(\mathcal{F}_c\) is the set of input features of the reward model assigned to
component \(c\). SHAP value comes from auxiliary Random Forest model trained on the same 1,292 data as the reward model. The Components with higher SHAP importance in the auxiliary Random
Forest therefore contribute more to the similarity score.
These weights are not direct feature-importance estimates for
the TabPFN reward model. The \texttt{Pub\_Year}
feature is excluded because it is a metadata attribute rather than a recipe
property. Component weights are listed in
Table~\ref{tab:similarity_weights}.

\begin{table}[htbp]
    \centering
    \caption{Component weights for the similarity metric, derived by summing
    the SHAP feature importances of the separate random forest model within each component.
    The \texttt{Pub\_Year} feature is excluded as a metadata attribute
    rather than a recipe property. Weights are not normalized to sum to
    unity; the metric normalizes internally.}
    \label{tab:similarity_weights}
    \footnotesize
    \setlength{\tabcolsep}{4pt}
    \renewcommand{\arraystretch}{1.1}
    \begin{tabular*}{\textwidth}{@{\extracolsep{\fill}} l
        >{\raggedright\arraybackslash}p{7.2cm}
        c @{}}
        \toprule
        \textbf{Component} & \textbf{Feature(s)} & \textbf{Weight \(W_c\)} \\
        \midrule
        Composition            & \texttt{comp\_Cs}, \texttt{comp\_FA}, \texttt{comp\_MA},
                                 \texttt{comp\_Pb}, \texttt{comp\_Sn},
                                 \texttt{comp\_Br}, \texttt{comp\_I} & 0.309 \\
        Substrate stack        & \texttt{sub\_layer\_1}, \texttt{sub\_layer\_2} & 0.173 \\
        ETL stack              & \texttt{etl\_layer\_1}, \texttt{etl\_layer\_2},
                                 \texttt{etl\_layer\_3} & 0.139 \\
        HTL stack              & \texttt{htl\_layer\_1}, \texttt{htl\_layer\_2} & 0.113 \\
        Band gap               & \texttt{Perovskite\_band\_gap} & 0.092 \\
        ETL deposition         & \texttt{ETL\_deposition\_procedure} & 0.066 \\
        Perovskite deposition  & \texttt{Perovskite\_deposition\_procedure} & 0.053 \\
        Backcontact stack      & \texttt{back\_layer\_1}, \texttt{back\_layer\_2} & 0.052 \\
        Quenching
            & \texttt{Perovskite\_\allowbreak deposition\_\allowbreak
                     quenching\_\allowbreak induced\_\allowbreak
                     crystallization}
            & 0.019 \\
        Backcontact deposition & \texttt{Backcontact\_deposition\_procedure} & 0.012 \\
        HTL deposition         & \texttt{HTL\_deposition\_procedure} & 0.010 \\
        \bottomrule
    \end{tabular*}
\end{table}

Composition carries the largest total weight (0.309), driven
by comp\_FA (0.154) and comp\_MA (0.087). The substrate stack
(0.173) and ETL stack (0.139) are the next most influential,
followed by the HTL stack (0.113) and band gap (0.092). The
four deposition procedures and the quenching flag together
contribute approximately 0.16, indicating lower aggregate
SHAP importance in the auxiliary Random Forest than composition
and layer-stack features.

\subsection{Aggregation and missing-value handling}

Missing values are handled according to component type.
For layer stacks and deposition procedures, missing entries
are replaced by empty strings before TF-IDF transformation.
An empty string produces a zero vector and therefore has
zero cosine similarity to both empty and non-empty strings.
The corresponding component weight remains in the
aggregation denominator. Missing textual information
therefore reduces the overall similarity.

Missing reference band gaps are imputed from composition
during reference-pool preparation. If either band gap value
remains missing at comparison time, the band gap component
receives zero weight for that recipe pair. Similarly, if
either quenching flag is missing, the quenching component
receives zero weight. Missing quenching values are not
treated as reported False values. A textual component is omitted from the comparison if its
entire column is absent from either input data frame.

The aggregate similarity and zero-denominator
convention are defined in
Equation~\ref{eq:recipe_similarity} in the main text.
The component-specific rules below specify
\(m_c(q,r)\) and its interaction with missing values.

For textual components whose columns are present,
$m_c(q,r)=1$ even when an individual entry is missing;
the missing entry instead contributes zero component
similarity. For band gap and quenching,
$m_c(q,r)=0$ when either value is missing and is one
otherwise. For composition, the component score and
weight are obtained from the element-wise SHAP-weighted
similarities described above.

Thus, the denominator can vary between recipe pairs
because missing numerical and Boolean components are
excluded, whereas missing textual entries retain their
nominal weights. The resulting similarity lies in
$[0,1]$, but a record with missing textual information
need not have self-similarity equal to one.

The full query--reference similarity matrix has shape
$(N_q,N_r)$. Neighborhoods are extracted from this matrix
using the protocol described in
Section~\ref{sec:neighborhood_methods}.

\subsection{Properties and limitations}

\paragraph{Why SHAP weighting.}
A uniformly weighted similarity over all feature groups would give equal importance to components with different predictive contributions in the auxiliary Random Forest. SHAP weighting instead emphasizes components with higher feature importance in that model. This provides a model-informed weighting of recipe resemblance, but does not establish which components the TabPFN reward model relies on most.

\paragraph{Symmetric SMAPE for composition.}
For each species \(e \in \{\mathrm{Cs},\mathrm{FA},\mathrm{MA},\mathrm{Pb},
\mathrm{Sn},\mathrm{Br},\mathrm{I}\}\) the element-wise score is
\begin{equation}
s_e^{\mathrm{comp}}(q,r) \;=\; 1 - \frac{\bigl|\hat{x}_e(q) - \hat{x}_e(r)\bigr|}
{\bigl|\hat{x}_e(q)\bigr| + \bigl|\hat{x}_e(r)\bigr| + \epsilon},
\end{equation}
with \(\epsilon = 10^{-9}\), where \(\hat{x}_e\) is the fraction of species
\(e\) after normalizing the seven parsed counts so that
\(\sum_e \hat{x}_e = 1\). The score is symmetric under swapping \(q\) and
\(r\) and bounded in \([0,1]\). It is well defined at zero fractions: a
species present in one composition and absent in the other receives
\(s_e^{\mathrm{comp}} = 0\), and a species absent from both receives
\(s_e^{\mathrm{comp}} = 1\). For example, comparing pure-\(\mathrm{MAPbI_3}\)
with pure-\(\mathrm{FAPbI_3}\) gives
\(\hat{x} = (0.2,0,0.2,0.6)\) for (MA, FA, Pb, I) versus
\(\hat{x} = (0,0.2,0.2,0.6)\), so the element-wise scores are \(0\) for MA
and FA, \(1\) for the shared Pb and I blocks, and \(1\) for the jointly
absent Cs, Sn, and Br. The composition component \(S_{\mathrm{comp}}(q,r)\)
is the SHAP-weighted average of these seven element-wise scores, and is
therefore not zero for this pair: the differing A-site species are
penalized element-wise, but the shared Pb/I framework and the jointly
absent species still contribute. This behaviour is preferable to cosine
similarity on the raw composition vector, which does not separately
score the substituted A-site species: for the same pair cosine gives
\(\cos \approx 10/11 \approx 0.91\), because the shared Pb and I entries
dominate the dot product and the norm. In the implementation, this
SHAP-weighted average over the seven species is applied inside the
aggregate of Eq.~\eqref{eq:recipe_similarity} rather than folded into a single
composition weight; the two forms are algebraically identical, since the
per-element weights sum to the composition weight reported in
Table~\ref{tab:similarity_weights}.

\paragraph{Limitations.}
The metric does not account for precursor identity, solvent choices,
temperature profiles, or layer thicknesses. These exclusions reflect limitations in data availability. The Perovskite
Database reports these attributes inconsistently across source papers, with
substantial missingness in the fields that would be required to encode them
as reliable features. Training the reward model on such sparse columns would
introduce more noise than signal, so they were excluded from the feature set
and, consequently, from the similarity metric. A generated recipe matching a
real device under our metric may therefore still differ from it in synthesis
details that are not represented in the database. The metric is a filter for
compositional and architectural similarity, not a proof of synthesizability,
and experimental validation of the top-ranked recipes remains the necessary
next step, as discussed in Section~\ref{sec:discussion}. Within the reference pool, the metric's neighborhood signal
is strongly rank-correlated with measured PCE
(Spearman $\rho \approx 0.70$; see
Appendix~\ref{app:signal_validation}). This association
supports the informativeness of the metric within that pool,
despite the limitations of its compositional and
architectural scope.

\subsection{Supported Hit Rate signal analysis}
\label{app:coverage-adjusted}

\begin{figure}[htbp]
    \centering
    \includegraphics[width=\textwidth]
    {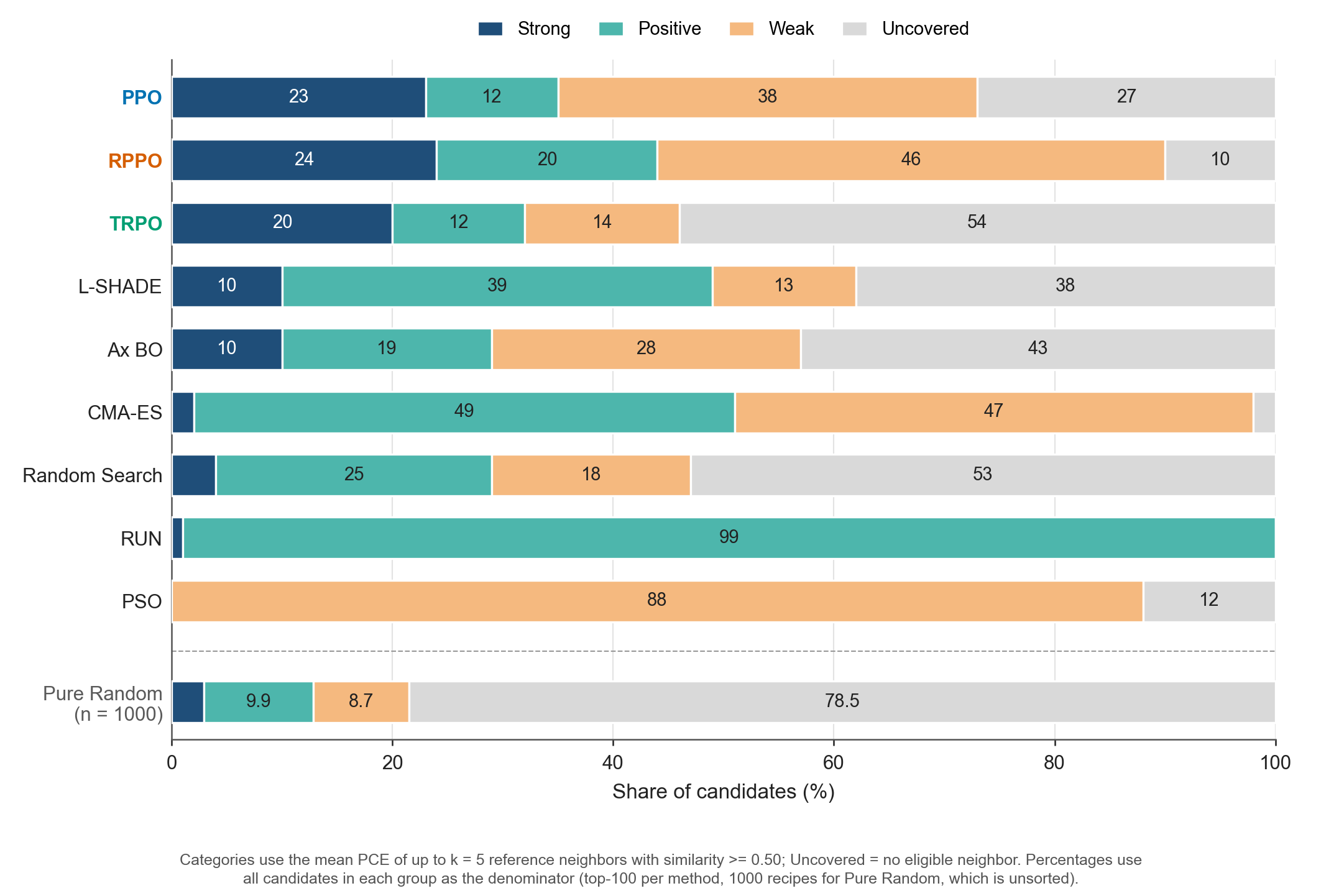}
    \caption{
    Reference-neighborhood evidence across all compared
    optimization methods.
    Bars show the proportions classified as Strong, Positive, Weak,
    or Uncovered within each method's top-100 surrogate-ranked
    shortlist, extending Figure~\ref{fig:neighbourhood} to all methods
    included in the candidate-level analysis: the three
    reinforcement-learning policies (PPO, RPPO, and TRPO) and six
    optimization baselines (L-SHADE, Ax BO, CMA-ES, Random Search,
    RUN, and PSO).
    The lower bar shows Pure Random, a background reference of
    1{,}000 unselected recipes that were not surrogate-ranked; it is
    not a matched-shortlist comparison and should not be confused
    with the Random Search baseline.
    All groups use the same reference pool, a maximum of $k=5$
    neighbors, and a minimum similarity of $0.50$.
    Neighborhood PCE is the unweighted mean measured PCE of the
    eligible neighbors. Strong denotes neighborhood PCE at or above
    the reference 85th percentile; Positive denotes values at or
    above the reference median but below the 85th percentile; Weak
    denotes values below the median. Candidates without an eligible
    neighbor are Uncovered and are not classified as Weak.
    Percentages use all candidates in each group as the denominator
    (100 per method; 1{,}000 for Pure Random). Segments below 7\% are
    not labeled.
    The PSO and RUN shortlists are not diverse in reference support:
    all 88 covered PSO candidates matched the same single reference
    device (PCE 17.90\%), and RUN neighborhoods were drawn from a
    few reference devices, so their category fractions reflect a
    narrow region of the design space rather than independent
    evidence from many candidates.
    These categories describe reference-neighborhood evidence, not
    experimentally verified candidate performance.
    }
    \label{fig:neighbourhood_all_methods}
\end{figure}

Figure~\ref{fig:neighbourhood_all_methods} extends the
main-text neighborhood-support comparison to all evaluated
optimization methods. Bars show Strong, Positive, Weak,
and Uncovered fractions using each complete top-100 shortlist
as the denominator.

The Strong percentages in Table~\ref{tab:signal_distribution}
are conditional on having at least one eligible reference neighbor.
To account for differences in coverage, we additionally calculated
the fraction of all generated recipes that were both covered
and classified as Strong:
\begin{equation}
F_{\mathrm{C},S_k} = C \times S_k,
\label{eq:covered-strong}
\end{equation}
where \(C\) is the coverage fraction and \(S_k\) is the
Strong fraction among covered recipes, with both
expressed as proportions.
Coverage is common to all reported \(k\), because eligibility
requires at least one reference neighbor satisfying the same
similarity threshold.

\begin{table}[htbp]
    \centering
    \caption{Signal distribution of generated recipes for each method.
    For each recipe, up to \(k\) nearest reference recipes with
    similarity \(\geq\tau=0.50\) are selected from the deduplicated
    Perovskite Database and LLM-extracted reference pool (\(n=890\); see
    Appendix~\ref{app:data:prep}); their measured PCE values define
    the neighborhood PCE. Recipes without an eligible neighbor are
    \textit{uncovered}. Covered recipes are labeled \textbf{Strong}
    (\(\geq\)Q85), \textbf{Pos.} (\(\geq\)Q50 and \(<\)Q85), or
    \textbf{Weak} (\(<\)Q50) by neighborhood PCE.
    \textbf{Avg. PCE} is the mean\,$\pm$\,std of recipe-level
    neighborhood PCE (\%) among covered recipes. \textbf{Cov.} is the
    fraction of analyzed candidates with at least one eligible
    neighbor (shared across \(k\)). Strong/Pos./Weak percentages use
    covered recipes as denominator; Cov.\ uses the full analyzed set
    (100 shortlisted candidates per RL policy or baseline; 1{,}000
    unselected samples for Pure Random). Percentages may not sum to
    100 due to rounding.}
    \label{tab:signal_distribution}
    \footnotesize
    \setlength{\tabcolsep}{4pt}
    \renewcommand{\arraystretch}{1.05}

    \begin{tabular*}{\columnwidth}{@{\extracolsep{\fill}}l cccc c@{}}
    \toprule
    \multicolumn{6}{c}{\textbf{k = 3}} \\
    \cmidrule(lr){1-6}
    \textbf{Method} & \textbf{Strong} & \textbf{Pos.} & \textbf{Weak}
    & \textbf{Avg. PCE} & \textbf{Cov.} \\
    \midrule
    \multicolumn{6}{l}{\textit{Reinforcement learning agents}} \\
    PPO   & 31.5\% & 21.9\% & 46.6\% & 18.64 ± 4.36\% & 73.0\% \\
    RPPO  & 26.7\% & 25.6\% & 47.8\% & 18.23 ± 4.39\% & 90.0\% \\
    TRPO  & 45.7\% & 34.8\% & 19.6\% & 20.65 ± 3.73\% & 46.0\% \\
    \midrule
    \multicolumn{6}{l}{\textit{Optimization baselines and sampling reference}} \\
    L-SHADE       & 16.1\% & 69.4\% & 14.5\% & 20.38 ± 2.43\% & 62.0\% \\
    Ax BO         & 17.5\% & 28.1\% & 54.4\% & 18.09 ± 4.39\% & 57.0\% \\
    Random Search &  6.4\% & 55.3\% & 38.3\% & 18.44 ± 3.60\% & 47.0\% \\
    CMA-ES        &  3.1\% & 56.1\% & 40.8\% & 18.36 ± 3.32\% & 98.0\% \\
    RUN           &  1.0\% & 99.0\% &  0.0\% & 21.18 ± 0.27\% & 100.0\% \\
    PSO           &  0.0\% &  0.0\% & 100.0\% & 17.90 ± 0.00\% & 88.0\% \\
    Pure Random   & 14.9\% & 47.0\% & 38.1\% & 18.35 ± 4.27\% & 21.5\% \\
    \bottomrule
    \end{tabular*}
    
    \par\vspace{1.2ex}
    
    \begin{tabular*}{\columnwidth}{@{\extracolsep{\fill}}l cccc c@{}}
    \toprule
    \multicolumn{6}{c}{\textbf{k = 5}} \\
    \cmidrule(lr){1-6}
    \textbf{Method} & \textbf{Strong} & \textbf{Pos.} & \textbf{Weak}
    & \textbf{Avg. PCE} & \textbf{Cov.} \\
    \midrule
    \multicolumn{6}{l}{\textit{Reinforcement learning agents}} \\
    PPO   & 31.5\% & 16.4\% & 52.1\% & 19.01 ± 3.72\% & 73.0\% \\
    RPPO  & 26.7\% & 22.2\% & 51.1\% & 18.70 ± 3.54\% & 90.0\% \\
    TRPO  & 43.5\% & 26.1\% & 30.4\% & 20.33 ± 3.61\% & 46.0\% \\
    \midrule
    \multicolumn{6}{l}{\textit{Optimization baselines and sampling reference}} \\
    L-SHADE       & 16.1\% & 62.9\% & 21.0\% & 20.35 ± 2.17\% & 62.0\% \\
    Ax BO         & 17.5\% & 33.3\% & 49.1\% & 18.30 ± 4.24\% & 57.0\% \\
    Random Search &  8.5\% & 53.2\% & 38.3\% & 18.50 ± 3.43\% & 47.0\% \\
    CMA-ES        &  2.0\% & 50.0\% & 48.0\% & 18.33 ± 2.62\% & 98.0\% \\
    RUN           &  1.0\% & 99.0\% &  0.0\% & 21.60 ± 0.19\% & 100.0\% \\
    PSO           &  0.0\% &  0.0\% & 100.0\% & 17.90 ± 0.00\% & 88.0\% \\
    Pure Random   & 13.5\% & 46.0\% & 40.5\% & 18.35 ± 4.17\% & 21.5\% \\
    \bottomrule
    \end{tabular*}
    
    \par\vspace{1.2ex}
    
    \begin{tabular*}{\columnwidth}{@{\extracolsep{\fill}}l cccc c@{}}
    \toprule
    \multicolumn{6}{c}{\textbf{k = 10}} \\
    \cmidrule(lr){1-6}
    \textbf{Method} & \textbf{Strong} & \textbf{Pos.} & \textbf{Weak}
    & \textbf{Avg. PCE} & \textbf{Cov.} \\
    \midrule
    \multicolumn{6}{l}{\textit{Reinforcement learning agents}} \\
    PPO   & 31.5\% & 28.8\% & 39.7\% & 19.68 ± 3.25\% & 73.0\% \\
    RPPO  & 26.7\% & 34.4\% & 38.9\% & 19.41 ± 2.97\% & 90.0\% \\
    TRPO  & 43.5\% & 30.4\% & 26.1\% & 20.61 ± 3.33\% & 46.0\% \\
    \midrule
    \multicolumn{6}{l}{\textit{Optimization baselines and sampling reference}} \\
    L-SHADE       & 14.5\% & 77.4\% &  8.1\% & 20.58 ± 1.97\% & 62.0\% \\
    Ax BO         & 15.8\% & 36.8\% & 47.4\% & 18.25 ± 4.16\% & 57.0\% \\
    Random Search &  6.4\% & 53.2\% & 40.4\% & 18.57 ± 3.21\% & 47.0\% \\
    CMA-ES        &  1.0\% & 60.2\% & 38.8\% & 18.72 ± 1.80\% & 98.0\% \\
    RUN           &  1.0\% & 99.0\% &  0.0\% & 21.60 ± 0.19\% & 100.0\% \\
    PSO           &  0.0\% &  0.0\% & 100.0\% & 17.90 ± 0.00\% & 88.0\% \\
    Pure Random   & 13.5\% & 46.5\% & 40.0\% & 18.30 ± 4.14\% & 21.5\% \\
    \bottomrule
    \end{tabular*}
    
    \par\smallskip
    \noindent\textit{All 88 covered PSO candidates were classified as Weak
    with identical neighborhood PCE (17.90\%) at every $k$.
    No RUN candidate had more than five eligible neighbors (four had
    exactly five), so its $k = 5$ and $k = 10$ values coincide.}
\end{table}

Table~\ref{tab:coverage_adjusted_strong} reports the
percentages of all candidates classified as both covered
and Strong, using the full candidate-set denominator.
RPPO achieved the highest fraction at each reported
maximum neighborhood size (24.0\%), followed by PPO
(23.0\%). TRPO reached 21.0\% at $k = 3$ and 20.0\%
at both $k = 5$ and $k = 10$, exceeding the corresponding
L-SHADE values of 10.0\%, 10.0\%, and 9.0\% (tied with Ax BO).
All three RL methods exceeded the evaluated non-RL
baselines on this metric at each reported maximum
neighborhood size.
This measure quantifies database-supported Strong
classifications; it does not imply that uncovered
recipes have low PCE or establish the measured
performance of generated recipes.

\begin{table}[htbp]
    \centering
    \caption{
    Percentages of analyzed candidates that are both covered
    and classified as Strong, calculated as the product of
    the coverage fraction and the conditional Strong
    fraction from Table~\ref{tab:signal_distribution}.
    The denominator is 100 shortlisted candidates for each
    RL policy and optimization baseline, and 1,000
    unselected samples for Pure Random. Pure Random
    characterizes background support under sampling
    without surrogate-based selection and is not a
    matched-shortlist comparison. The parameter \(k\)
    denotes the maximum neighborhood size. Uncovered
    candidates are not classified as Weak. PSO has a zero supported hit rate because all 88 of its covered candidates were classified as Weak.
    }
    \label{tab:coverage_adjusted_strong}
    \begin{tabular*}{\textwidth}{@{\extracolsep{\fill}}lrrr@{}}
        \toprule
        Method & $k = 3$ & $k = 5$ & $k = 10$ \\
        \midrule
        PPO           & 23.0\% & 23.0\% & 23.0\% \\
        RPPO          & 24.0\% & 24.0\% & 24.0\% \\
        TRPO          & 21.0\% & 20.0\% & 20.0\% \\
        \midrule
        L-SHADE       & 10.0\% & 10.0\% &  9.0\% \\
        Ax BO         & 10.0\% & 10.0\% &  9.0\% \\
        Random Search &  3.0\% &  4.0\% &  3.0\% \\
        RUN           &  1.0\% &  1.0\% &  1.0\% \\
        CMA-ES        &  3.0\% &  2.0\% &  1.0\% \\
        PSO           &  0.0\% &  0.0\% &  0.0\% \\
        Pure Random   &  3.2\% &  2.9\% &  2.9\% \\
        \bottomrule
    \end{tabular*}
\end{table}

\subsection{Rank-resolved neighborhood support}
\label{app:early-enrichment}

\begin{figure}[htbp]
    \centering
    \includegraphics[width=\textwidth]
    {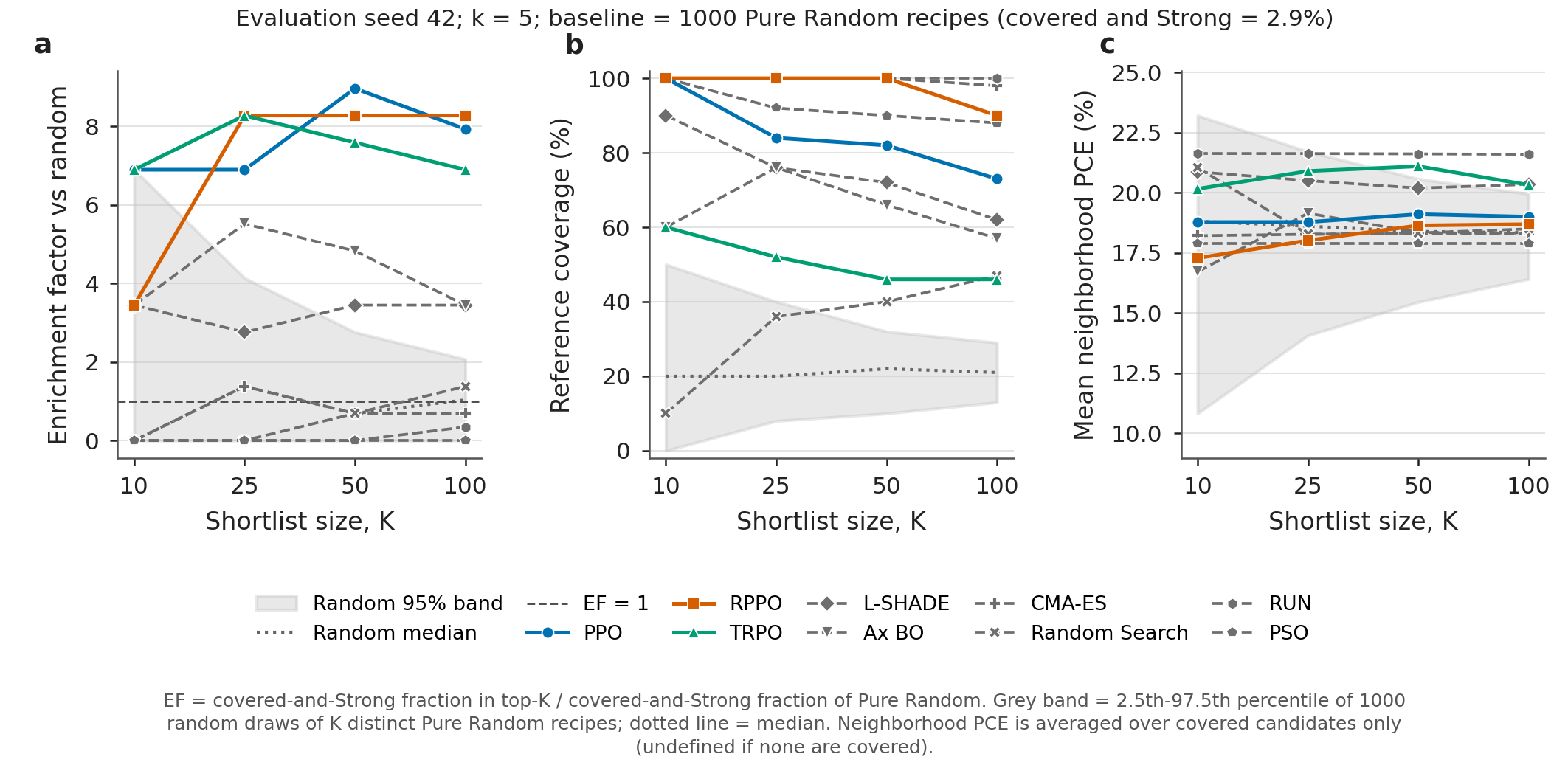}
    \caption{
    Early enrichment and screening-quality trade-offs across all
    evaluated optimization methods.
    This figure extends Figure~\ref{fig:early_enrichment} to
    the three reinforcement-learning policies (PPO, RPPO, and
    TRPO; colored solid lines) and six optimization baselines
    (L-SHADE, Ax BO, CMA-ES, Random Search, RUN, and PSO; grey
    dashed lines, distinguished by marker). Candidates are
    ranked by descending saved surrogate-predicted PCE, and
    statistics are evaluated for top-$K$ prefixes at $K=10$,
    25, 50, and 100.
    (a) Enrichment factor (EF): the neighborhood-supported hit
    rate (the fraction of all $K$ candidates that are covered
    and have mean neighborhood PCE at or above the
    reference-pool 85th percentile) divided by the same fraction
    among the 1{,}000 unselected Pure Random recipes (2.9\% at
    $k=5$). The dashed horizontal line marks $\mathrm{EF}=1$.
    Candidates without a reference neighbor count as non-hits.
    (b) Reference coverage: the fraction of all $K$ candidates
    with at least one eligible reference neighbor.
    (c) Mean candidate-level neighborhood PCE among covered
    candidates only; it is undefined when no candidate in the
    prefix is covered.
    In all panels, the grey band spans the 2.5th--97.5th
    percentiles of the statistic over 1{,}000 random draws of
    $K$ distinct recipes from the Pure Random sample, and the
    dotted line is the median of those draws.
    Pure Random is not plotted as a method because its samples
    were not ranked by surrogate score.
    For each candidate, neighborhood PCE is the unweighted mean
    measured PCE of up to $k=5$ reference neighbors with
    similarity $\geq0.50$. $K$ denotes shortlist size and $k$
    the maximum number of reference neighbors.
    All methods use the same reference pool and classification
    thresholds, and results use the designated seed-42 run for
    each method. Lines connect evaluated shortlist sizes as
    visual guides only; they do not represent optimization
    trajectories or uncertainty across runs, and the band
    reflects sampling variability of unselected selection only.
    These statistics describe experimental-reference support,
    not measured performance of generated devices.
    }
    \label{fig:early_enrichment_all_methods}
\end{figure}

Figure~\ref{fig:early_enrichment_all_methods} extends the
main-text rank-resolved analysis to all evaluated optimization
methods. Neighborhood-supported hit rates, reference coverage,
and mean neighborhood PCE among covered candidates are
evaluated for top-$K$ prefixes at $K=10$, 25, 50, and 100,
using the designated seed-42 shortlists, $k=5$, and the
same reference pool and similarity threshold.
Panel (a) expresses the supported hit rate as an
enrichment factor relative to the Pure Random recipes
(Equation~\ref{eq:enrichment_factor}). Pure Random is not
plotted as a method, because its samples were not ordered
by surrogate score; it supplies only the enrichment-factor
denominator and the grey random-selection interval. It
should not be confused with the Random Search baseline, an
optimization method whose candidates are surrogate-ranked
like those of the other methods.

\subsection{Validation of the Neighborhood Signal}
\label{app:signal_validation}

The neighborhood analysis characterizes experimental
reference support for generated candidates using
the mean measured PCE of similar reference devices.
To assess whether this criterion contains
performance-related information within the
reference pool, we applied the same top-$k$
procedure to each reference recipe, excluded
self-matches, and retained the similarity
threshold $\tau=0.50$. We then correlated
neighborhood-mean PCE with the reference recipe's
own measured PCE. This internal assessment does
not directly validate generated-device performance.

Table~\ref{tab:signal_validation} reports the results for
$k \in \{3, 5, 10\}$. Coverage is $96.4\%$ in every case, so the analysis is
not restricted to a small subpopulation. Spearman $\rho$ ranges from $+0.682$
to $+0.707$, with narrow bootstrap confidence intervals. The observed
$\rho$ exceeds the mean of the PCE-shuffled null
distribution by more than 16 standard deviations
($p_{\mathrm{perm}} = 0.002$, the minimum resolvable
permutation $p$-value). This supports an association between the similarity
graph and measured PCE under the specified
permutation control. Interpretation depends on
how the supervised similarity weights were fitted
and on dependence among reference records.

We additionally tested whether \emph{top-$k$ selection}
contributes beyond what random neighbors in the same
pool would provide. We used a neighbor-shuffle null
that replaces each recipe's top-$k$ neighbors with
$k$ reference recipes drawn uniformly at random,
keeping the measured PCE labels fixed. The mean null
correlation, $\bar\rho_\text{nbr-shuf}$, was close to
zero at every evaluated $k$ (within $\pm 0.005$).
Consequently, the increment
$\Delta\rho = \rho - \bar\rho_\text{nbr-shuf}$
was approximately equal to the observed $\rho$.
These results support the contribution of
similarity-based neighbor selection to the observed
association. Together, the controls support the
informativeness of neighborhood scores within the
reference pool, but do not establish predictive
accuracy for optimizer-generated candidates.

\begin{table}[h]
\centering
\small
\caption{Validation of the SHAP-weighted similarity metric on the reference
pool ($n = 890$). For each reference recipe, we compute the mean PCE of its
top-$k$ most similar neighbors and correlate this neighborhood score with the
recipe's own measured PCE. Neighbors are filtered by a similarity floor
$\tau = 0.50$; self-matches are excluded. \textbf{Cov.} is the fraction of
recipes with $\geq 1$ eligible neighbor. $\rho$ is Spearman's rank correlation
with a bootstrap $95\%$ CI ($N_\text{boot} = 1000$).
\textbf{$Z_\text{PCE-shuf}$} is the number of standard deviations by which the
observed $\rho$ exceeds a PCE-shuffled null that holds the similarity graph
fixed and permutes PCE labels ($N_\text{perm} = 500$; $p_\text{perm} = 0.002$
in every case). \textbf{$\bar\rho_\text{nbr-shuf}$} is the mean $\rho$ under a
neighbor-shuffle null that replaces the top-$k$ neighbors with $k$ recipes
drawn uniformly at random. \textbf{$\Delta\rho$} is the increment of the
observed $\rho$ over this null ($\Delta\rho = \rho - \bar\rho_\text{nbr-shuf}$),
isolating the contribution of similarity-based selection from pool-wide PCE
autocorrelation. All entries are rounded to three decimals; $\Delta\rho$ is
computed from the unrounded $\rho$ and $\bar\rho_\text{nbr-shuf}$, so in some
rows the displayed values may differ in the last digit from
$\rho - \bar\rho_\text{nbr-shuf}$ computed on the rounded entries.}
\label{tab:signal_validation}
\begin{tabular*}{\textwidth}{@{\extracolsep{\fill}}lcccccc@{}}
\toprule
$k$ & Cov. & $\rho$ & $95\%$ CI & $Z_\text{PCE-shuf}$ &
$\bar\rho_\text{nbr-shuf}$ & $\Delta\rho$ \\
\midrule
3  & 96.4\% & $+0.682$ & [$+0.636$, $+0.722$] & $+17.55$ & $-0.001$ & $+0.683$ \\
5  & 96.4\% & $+0.699$ & [$+0.655$, $+0.740$] & $+17.38$ & $-0.004$ & $+0.702$ \\
10 & 96.4\% & $+0.707$ & [$+0.664$, $+0.742$] & $+16.81$ & $-0.004$ & $+0.711$ \\
\bottomrule
\end{tabular*}
\end{table}

\section{OpenAlex Retrieval and Composition-Family Matching}
\label{app:openalex}

\subsection{Search queries}
\label{app:openalex-queries}

The following eleven search terms were used to retrieve candidate
papers from OpenAlex~\cite{priem2022openalex}. Each query was restricted to non-open-access works published
from 2024 onwards, and duplicate works across queries were removed by
OpenAlex ID before extraction.
\begin{itemize}
    \item \texttt{perovskite solar cell composition}
    \item \texttt{tin-lead perovskite solar cell}
    \item \texttt{Sn-Pb perovskite solar cell}
    \item \texttt{all-inorganic perovskite solar cell}
    \item \texttt{CsPbI3 perovskite solar cell}
    \item \texttt{FAPbI3 perovskite solar cell}
    \item \texttt{mixed halide perovskite solar cell}
    \item \texttt{CsFA perovskite solar cell}
    \item \texttt{triple cation perovskite solar cell}
    \item \texttt{wide bandgap perovskite solar cell}
    \item \texttt{narrow bandgap perovskite solar cell}
\end{itemize}

OpenAlex exposes bibliographic metadata and, for a subset of works, an
abstract reconstructed from the \texttt{abstract\_inverted\_index}
field. For closed-access works, the full text is not available, and
layer- and deposition-level information that is reported in the methods
section rather than the abstract is therefore not recoverable. This is
the reason the external validation analysis is restricted to
composition. The perovskite composition is typically stated in the title
or the first sentence of the abstract, making it the most
reliably extractable field for this external comparison.

\subsection{Threshold sensitivity}
\label{app:openalex-threshold}

Table~\ref{tab:openalex-threshold} reports the composition-family matching rate as a
function of $\tau_\mathrm{comp}$. The rate is $99.5\%$ at
$\tau_\mathrm{comp}=0.50$ and $98.3\%$ at $0.60$, then falls to
$92.2\%$ at $0.70$, $65.4\%$ at the selected threshold of $0.80$, 
$35.8\%$ at $0.85$, $11.7\%$ at $0.90$, and $1.6\%$ at $0.95$.
The decline steepens between $0.80$ and $0.90$: the $53.7\%$ of
candidates whose best-match similarity lies in $[0.80,0.90)$
account for most of the drop across this range, whereas the
$26.8\%$ in $[0.70,0.80)$ account for the more moderate decline
between $0.70$ and $0.80$. These intervals describe the composition-similarity score only. As discussed
in Appendix~\ref{app:openalex-families}, the score does not map onto fixed chemical categories, so we do not assign a specific structural meaning (for example, a shared A/B/X sublattice framework) to any interval. The rationale for the threshold selected is given in the Section~\ref{sec:openalex-method}.

\begin{table}[t]
\centering
\caption{Distribution of best-match composition similarity
between $N = 9{,}000$ generated recipes and the retrieved
OpenAlex reference pool. A total of 1{,}054 recipes
(11.7\%) attained a best-match composition similarity of
0.90 or above, and no generated recipe exceeded a
similarity of 1.00. At the threshold of
$\tau_{\mathrm{comp}} = 0.80$, 5{,}889 recipes (65.4\%) 
were classified as composition-family matches. These
results describe similarity to the retrieved reference
compositions, not experimental validation or novelty
relative to the complete literature.}
\label{tab:openalex-distribution}
\begin{tabular*}{\columnwidth}{@{\extracolsep{\fill}}lrr@{}}
\toprule
Similarity interval & Recipes & Fraction \\
\midrule
$[0.0, 0.3)$  &     0 & $0.0\%$ \\
$[0.3, 0.5)$  &    48 & $0.5\%$ \\
$[0.5, 0.7)$  &   654 & $7.3\%$ \\
$[0.7, 0.8)$  & 2{,}409 & $26.8\%$ \\
$[0.8, 0.9)$  & 4{,}835 & $53.7\%$ \\
$[0.9, 1.0]$  & 1{,}054 & $11.7\%$ \\
\midrule
Total         & 9{,}000 & $100.0\%$ \\
\bottomrule
\end{tabular*}
\end{table}

A substantial fraction of generated compositions closely
match compositions in the retrieved OpenAlex reference pool
under the chosen metric: $35.8\%$ meet the $0.85$ threshold,
$11.7\%$ meet $0.90$, and $1.6\%$ meet $0.95$. Under this metric, few generated compositions are near-identical to a
retrieved composition ($1.6\%$ at $\tau_{\mathrm{comp}} = 0.95$), while most
are moderately similar to at least one ($92.2\%$ at $0.70$). However, these results do not
establish that every generated composition is novel relative
to the full published literature, nor do they identify the
mechanism by which the policies explore stoichiometric ratios.

At the selected threshold of $\tau_{\mathrm{comp}} = 0.80$, a majority of
generated recipes ($5{,}889$ of $9{,}000$; $65.4\%$) match a reference
composition family. The policy-specific rates are $74.9\%$ for RPPO,
$69.5\%$ for PPO, and $51.9\%$ for TRPO
(Table~\ref{tab:openalex-per-algorithm}). All three exceed the
$37.7\%$ rate of the unselected Pure Random background. In the designated
seed-42 runs, the ordering RPPO $>$ PPO $>$ TRPO holds at all seven
evaluated thresholds ($0.50$--$0.95$), and all three policies exceed the
Pure Random background at each of them
(Table~\ref{tab:openalex-threshold}). At $\tau_{\mathrm{comp}} = 0.50$ and
$0.60$, the differences among the policies are small (under $3$~pp). Because
each policy is represented by a single seed-42 candidate pool, this ordering
is not evidence of robustness across training seeds. We make no claim about
the ordering at unevaluated thresholds, and we do not infer matching rates at
unreported thresholds by interpolation. External validation here concerns
composition-family agreement with the retrieved literature, not reproduction
of a complete published recipe or experimental validation of device
performance.

\begin{table}[t]
\centering
    \caption{Composition-family matching rate as a function of the similarity
    threshold $\tau_{\mathrm{comp}}$, shown as matched recipes (rate in \%) for
    each generation algorithm ($3{,}000$ recipes each), the pooled RL-generated
    recipes ($9{,}000$), and the unselected Pure Random background
    ($1{,}000$). In the designated seed-42 runs, the ordering
    RPPO $>$ PPO $>$ TRPO holds at every evaluated threshold, and all three
    algorithms exceed the Pure Random background. The pooled decline is steepest
    between $0.80$ and $0.90$, and the pooled difference from Pure Random is
    largest at the threshold of $0.80$ ($27.7$~pp). These results
    characterize matching against the retrieved reference pool, not experimental
    validation.}
    \label{tab:openalex-threshold}
    \footnotesize
    \setlength{\tabcolsep}{3pt}
    \begin{tabular*}{\columnwidth}{@{\extracolsep{\fill}}rrrrrr@{}}
    \toprule
    $\tau_{\mathrm{comp}}$ & RPPO & PPO & TRPO & Pooled RL & Pure Random \\
    \midrule
    0.50 & 2{,}995 (99.8) & 2{,}991 (99.7) & 2{,}966 (98.9) & 8{,}952 (99.5) &   979 (97.9) \\
    0.60 & 2{,}981 (99.4) & 2{,}970 (99.0) & 2{,}898 (96.6) & 8{,}849 (98.3) &   933 (93.3) \\
    0.70 & 2{,}866 (95.5) & 2{,}825 (94.2) & 2{,}607 (86.9) & 8{,}298 (92.2) &   773 (77.3) \\
    \textbf{0.80} & \textbf{2{,}246 (74.9)} & \textbf{2{,}085 (69.5)} & \textbf{1{,}558 (51.9)} & \textbf{5{,}889 (65.4)} & \textbf{377 (37.7)} \\
    0.85 & 1{,}309 (43.6) & 1{,}163 (38.8) &   750 (25.0) & 3{,}222 (35.8) &   159 (15.9) \\
    0.90 &   472 (15.7) &   402 (13.4) &   180 (6.0)  & 1{,}054 (11.7) &    32 (3.2) \\
    0.95 &    71 (2.4)  &    51 (1.7)  &    20 (0.7)  &   142 (1.6)  &     2 (0.2) \\
    \bottomrule
    \end{tabular*}
\end{table}

\begin{table}[t]
    \centering
    \caption{Composition-family matching rates by generation algorithm at the threshold $\tau_{\mathrm{comp}} = 0.80$. RPPO has the highest
    matching rate ($74.9\%$), followed by PPO ($69.5\%$) and TRPO ($51.9\%$);
    TRPO is $17.6$~pp below PPO and $23.0$~pp below RPPO. All three exceed the
    unselected Pure Random background ($37.7\%$). Each policy is represented by
    its designated seed-42 candidate pool, so these differences are not an
    estimate of training-seed variability. These rates do not establish
    differences in exploration breadth or experimental validity.}
    \label{tab:openalex-per-algorithm}
    \small
    \begin{tabular*}{\columnwidth}{@{\extracolsep{\fill}}lrrr@{}}
    \toprule
    Algorithm & Recipes & Matched & Rate \\
    \midrule
    RPPO & 3{,}000 & 2{,}246 & $74.9\%$ \\
    PPO  & 3{,}000 & 2{,}085 & $69.5\%$ \\
    TRPO & 3{,}000 & 1{,}558 & $51.9\%$ \\
    \midrule
    All RL (pooled) & 9{,}000 & 5{,}889 & $65.4\%$ \\
    Pure Random     & 1{,}000 &   377   & $37.7\%$ \\
    \bottomrule
    \end{tabular*}
\end{table}

\subsection{Interpretation of composition similarity and family matching}
\label{app:openalex-families}

The SMAPE-based composition similarity described in
Appendix~\ref{app:similarity} quantifies agreement in constituent
identities and stoichiometric ratios using SHAP-derived weights.
The score depends on the compositions being compared and the
selected weights; it does not provide a universal mapping between
numerical similarity and chemically defined composition families.
In particular, differences in one or more sublattices do not
necessarily correspond to fixed similarity-score ranges.

$\tau_{\mathrm{comp}} = 0.80$ is a threshold
(Section~\ref{sec:openalex-method}), not a universal chemical boundary
between families; Table~\ref{tab:openalex-threshold} shows how the matching rate
changes with this choice. A match
under this criterion indicates compositional resemblance to a
retrieved reference composition; it does not establish reproduction
of a complete published recipe, experimental viability, or device
performance.
\clearpage
\label{LastPage}

\end{document}